\documentclass[10pt,twocolumn,letterpaper]{article}

\usepackage{cvpr}      %

\usepackage{microtype}
\usepackage{tabularx}

\usepackage{xcolor}
\definecolor{mygreen}{HTML}{117733}
\usepackage{multirow}

\definecolor{tabfirst}{rgb}{0.76,0.93,0.71} %
\definecolor{tabsecond}{rgb}{0.98 , 0.93, 0.77} %
\definecolor{tabthird}{rgb}{1, 1, 0.7} %

\usepackage{graphicx}
\usepackage{adjustbox}

\definecolor{cvprblue}{rgb}{0.21,0.49,0.74}
\usepackage[pagebackref,breaklinks,colorlinks,allcolors=cvprblue]{hyperref}

\title{RoomLight: A 2.5D Illumination Prior for Indoor Environments}

\author{Andreea Ardelean \quad \quad \quad
Bernhard Egger\\
Friedrich-Alexander-Universität Erlangen-Nürnberg \\
{\tt\small \{andreea.ardelean, bernhard.egger\}@fau.de}
}

\begin{document}
\maketitle
\begin{abstract}
Ill-posed inverse problems require priors to constrain the solution space toward plausible outcomes. 
In inverse rendering, learned priors modeling the distribution of natural illumination improve the recovery of scene properties. 
However, existing models rely on the distant-illumination assumption, representing lighting as a far-field environment map. 
This limits their applicability to indoor scenes, where illumination is highly spatially varying due to finite-distance emitters, visibility changes, and parallax, all of which are poorly approximated by a single environment map. 
To address this, we introduce a spatially-aware illumination prior trained on real-world indoor panoramas and their estimated depth. 
Our variational autoencoder model learns a compact, optimizable latent space that decodes into HDR radiance and depth, parameterizing an area light emitter for direct integration into standard differentiable rendering pipelines.
This design bridges the plausibility guarantees of a learned prior with the gradient flow required for downstream optimization. 
Crucially, by jointly modeling radiance and depth, our prior captures the spatial structure of indoor illumination, instead of treating the light sources as infinitely distant. 
We demonstrate that this formulation enables spatially-varying illumination modeling and achieves higher-fidelity recovery of indoor lighting compared to existing approaches. 
Project page: \small{\url{https://andreead-a.github.io/RoomLight}
}
\end{abstract}
    
\section{Introduction}

\begin{figure}
    \centering
    \includegraphics[width=0.95\linewidth]{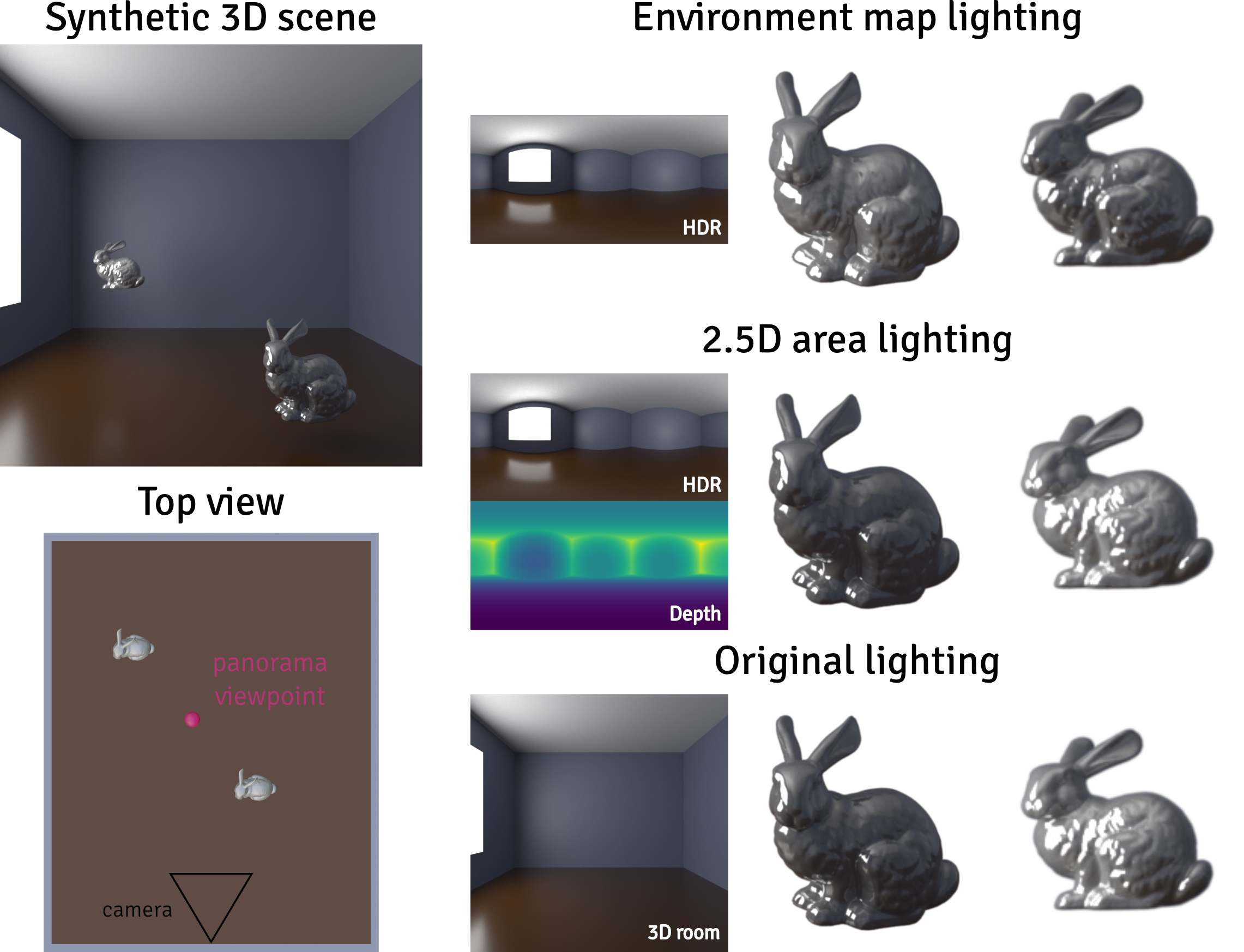}
    \caption{\label{fig:motivation_varying} Comparison of object relightings in an indoor environment. Conventional environment map lighting assumes distant illumination and cannot simultaneously reproduce the appearance of both objects. In contrast, a 2.5D area lighting representation, obtained by texturing a depth-based mesh with the HDR panorama, models spatially-varying illumination and closely reproduces the original rendering. Extended figure in the supplementary material.
    \vspace{-1.5em}}
\end{figure}

Recovering the illumination of a scene from input images with limited field of view (FOV) is a cornerstone of inverse rendering \cite{debevec98syn, barron2015shape, gardner2017indoor}, with applications including material estimation \cite{yu1999inverse, wu2023factorized}, scene relighting \cite{hasselgren2022shape, lin2025iris, careaga2025physically} and photorealistic object insertion \cite{garon2019fast, phongthawee2024diffusionlight}.
Yet, the problem is severely ill-posed as geometry, reflectance and lighting are tightly coupled in image formation.
Illumination estimation is particularly challenging because, at any surface point, incoming light arrives from almost every direction, often originating outside the camera's field of view, and spans a high dynamic range (HDR) of intensities. 

The dominant paradigm for representing scene illumination is the spherical environment map \cite{debevec98syn}, which encodes incident radiance as a function of direction at a single point.
Environment maps capture high-frequency lighting detail and therefore reproduce intense specular highlights and complex reflections on glossy surfaces faithfully, but they are spatially invariant by construction due to the distant-light assumption. 
While outdoor daylight scenes satisfy this assumption reasonably well, real interiors do not.
Nearby emitters produce strong parallax, and illumination varies significantly with position due to inverse-square falloff and changes in visibility, so a single environment map cannot match the underlying light field.

Intuitively, accounting for the finite distance between the illumination representation and the scene geometry removes this limitation.
Computer graphics and game engines have long exploited this through parallax-corrected cubemaps~\cite{parallax_cubemaps_2012}, which map the environment onto proxy geometry such as spheres~\cite{bjorke2004image} or boxes~\cite{valve_parallax_cubemaps}, but such proxies cannot represent real interior layouts.
We instead adopt a depth-based formulation: the HDR panorama is projected onto a mesh reconstructed from its own depth map, yielding what we call a \textit{2.5D area light}, also recently described as an \textit{emissive mesh}~\cite{manus2025depthlight}.
A single additional channel changes the representation class.
An environment map is a function of direction alone. Adding depth makes the emitter a surface in space, so the illumination it produces varies with position \emph{and} direction.
This allows illumination to vary naturally across the scene, so the same object appears differently depending on where it is placed.
Fig.~\ref{fig:motivation_varying} shows that even in a simple indoor scene, no single environment map can reproduce the appearance of two nearby objects simultaneously, regardless of capture position, whereas a 2.5D area light can.

2.5D area lights are directly compatible with existing differentiable rendering frameworks \cite{Mitsuba3}, but recovering one from a single image is even more underconstrained than recovering an environment map: the representation spans both radiance and geometry, and most of it is never observed.
Direct optimization is consequently prone to overfitting, often converging to noisy and implausible solutions.
Still, real-world illumination is far from
arbitrary, as indoor and outdoor environments alike exhibit strong statistical regularities \cite{dror2004statistical}.
Data-driven priors \cite{gardner2023reni++, gardner2022reni, walker2026veni, lyu2023diffpostillum, boss2021neural} exploit this property by replacing an unconstrained pixel grid with a compact latent space whose decoded samples remain plausible throughout optimization.
Existing priors, however, are direction-only by construction: their latent codes decode radiance as a function of direction alone, with no notion of how far the emitting surfaces are.

This leaves a gap: the representation that indoor scenes require has no prior, and existing priors cannot represent it.
We close this gap with a spatially-aware illumination prior tailored to indoor environments.
We train a convolutional variational autoencoder on equirectangular panoramas pairing HDR radiance with aligned depth, so a single latent code decodes both the emissive texture and the geometry of a 2.5D area light.
Decoding is differentiable end to end, letting photometric gradients from a differentiable renderer propagate back to the latent code.
The prior thus serves directly as the illumination parameterization in an analysis-by-synthesis setting for inverse rendering, where scene parameters are recovered by rendering a hypothesis and comparing it against the observation.
We show that optimizing in this latent space recovers spatially varying indoor illumination more accurately than environment-map priors fitted through the same renderer, and that it transfers to real photographs.

\begin{figure*}
    \centering
    \includegraphics[width=0.99\linewidth]{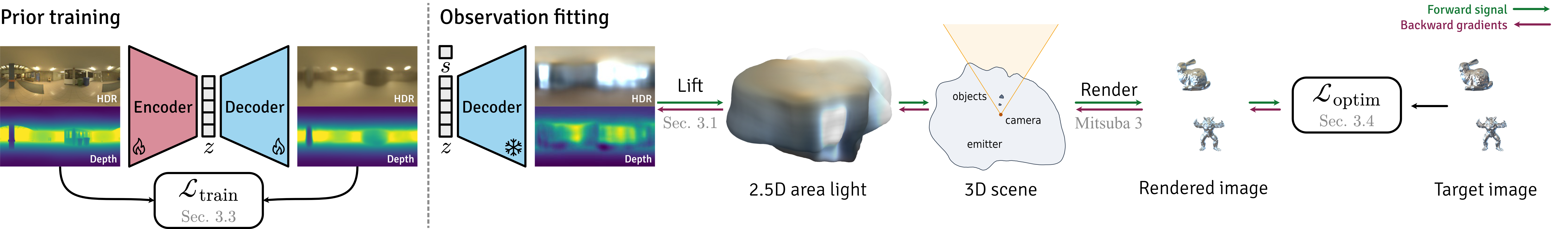}
    \caption{
    \label{fig:method}
    Method overview. We pre-train a convolutional VAE on paired HDR radiance and depth panoramas to encode an environment into a global latent vector $z$. For illumination recovery the encoder is discarded and the decoder is frozen. The decoded pair is lifted into a 2.5D area light which illuminates the objects in a differentiable renderer. The photometric loss between the rendered image and the target observation is backpropagated through all steps to the latent code $z$ and the exposure scale $s$.
    \vspace{-1em}} 
\end{figure*}

\section{Related work}

\paragraph{Lighting in inverse rendering.} 
Recovering intrinsic scene properties such as geometry, material reflectance and incident illumination from images is a long-standing problem in computer vision and graphics \cite{barrow1978recovering}.
Classical shape-from-shading formulations avoid the illumination unknown altogether, assuming a single distant source of known direction \cite{horn1970shapeshading}.
Virtually all methods that do estimate lighting retain the distant-illumination assumption and differ mainly in how they parameterize the resulting environment map \cite{debevec98syn}.
Low-order spherical harmonics (SH) \cite{ramamoorthi2001efficient, basri2003lambertian} and mixtures of spherical Gaussians (SG) \cite{tsai2006all, zhang2021physg} provide a compact, smooth basis that is cheap to optimize and adequate for diffuse reflectance, but their limited angular bandwidth cannot express the high-frequency illumination needed for sharp specular reflections.
Recent inverse-rendering pipelines therefore optimize the environment map non-parametrically, as an unconstrained pixel grid \cite{hasselgren2022shape, wang2025materialist}, trading compactness for expressiveness.
This optimization is driven by differentiable renderers \cite{Mitsuba3, Laine2020diffrast, redner2018} through photometric reconstruction losses, where a high-dimensional, unregularized lighting representation makes the problem severely under-constrained.
We target this trade-off with an indoor illumination prior that is expressive yet low-dimensional, and whose latent parameters are directly optimizable within such frameworks.

\vspace{-1em}
\paragraph{Illumination priors.}
Inverse rendering is inherently ill-posed, as multiple combinations of material and illumination can explain the same observed image.
Linear statistical models over parametric lighting representations capture distributions of low-dimensional illumination coefficients \cite{yu2021outdoor, egger2018occlusion, barron2015shape}, but are limited by the expressiveness of the underlying parametric model.
Learned environment map models are more expressive, ranging from convolutional autoencoders \cite{boss2021neural, sztrajman2021neural} to variational auto-decoders with neural field decoders \cite{gardner2022reni, gardner2023reni++} and large-scale variational autoencoders \cite{walker2026veni}.
Diffusion Posterior Illumination \cite{lyu2023diffpostillum} instead integrates a denoising diffusion model \cite{ho2020denoising} into optimization-based inverse rendering, improving realism and mitigating the material--illumination ambiguity.
Yet all these priors produce a single global estimate, ignoring the spatially varying nature of real indoor lighting.
We instead model the geometry of the illumination environment explicitly, enabling more accurate indoor inverse rendering.

\vspace{-1em}
\paragraph{Spatially-varying indoor illumination.} 
Early learning-based approaches to indoor illumination estimation predict a single global environment map for an entire scene \cite{gardner2017indoor, wang2022stylelight, dastjerdi2023everlight}, which fails to capture spatially varying lighting effects. 
To address this limitation, per-pixel lighting predictions using spatially-varying SH \cite{garon2019fast}, SG \cite{li2020inverse, zhu2022irisformer}, or locale-dependent HDR panoramas \cite{song2019neural} were introduced. 
However, these methods remain restricted to 2D image-space representations and do not enforce a coherent 3D lighting structure. 
\cite{gardner2019deep} directly predicts localized 3D light sources, but cannot capture the level of detail required for accurate reflections.
\cite{wang2021learning} introduced a volumetric SH representation defined on a voxel grid, enabling lighting queries at arbitrary 3D positions and view directions. 
This approach was later extended to video input in \cite{li2023spatiotemporally}. 
More recently, generative methods have demonstrated that spatially-varying illumination can be recovered by inpainting light probes in input images \cite{phongthawee2024diffusionlight} or video frames \cite{tong2025spatio4dsig, bolduc2025lighting}.
Alternatively, the limited FOV input images can be outpainted into full panoramas, which are then converted into emissive meshes using estimated panorama depth \cite{manus2025depthlight}.
While these methods achieve strong performance in their respective settings, they are primarily designed to directly predict illumination for specific applications such as object insertion. 
As a result, they do not provide a learned prior that can be incorporated into optimization-based inverse rendering frameworks. 
This limitation can lead to unrealistic illumination estimates and prevents the joint recovery of other scene properties such as material parameters.
Instead, our approach introduces a neural prior for spatially-varying illumination that can be integrated into inverse rendering pipelines, enabling more consistent estimation of lighting and other scene attributes.

\section{Method}

\begin{figure*}[ht]
    \centering
    \includegraphics[width=0.99\linewidth,     trim=0 0 0 1cm,clip]{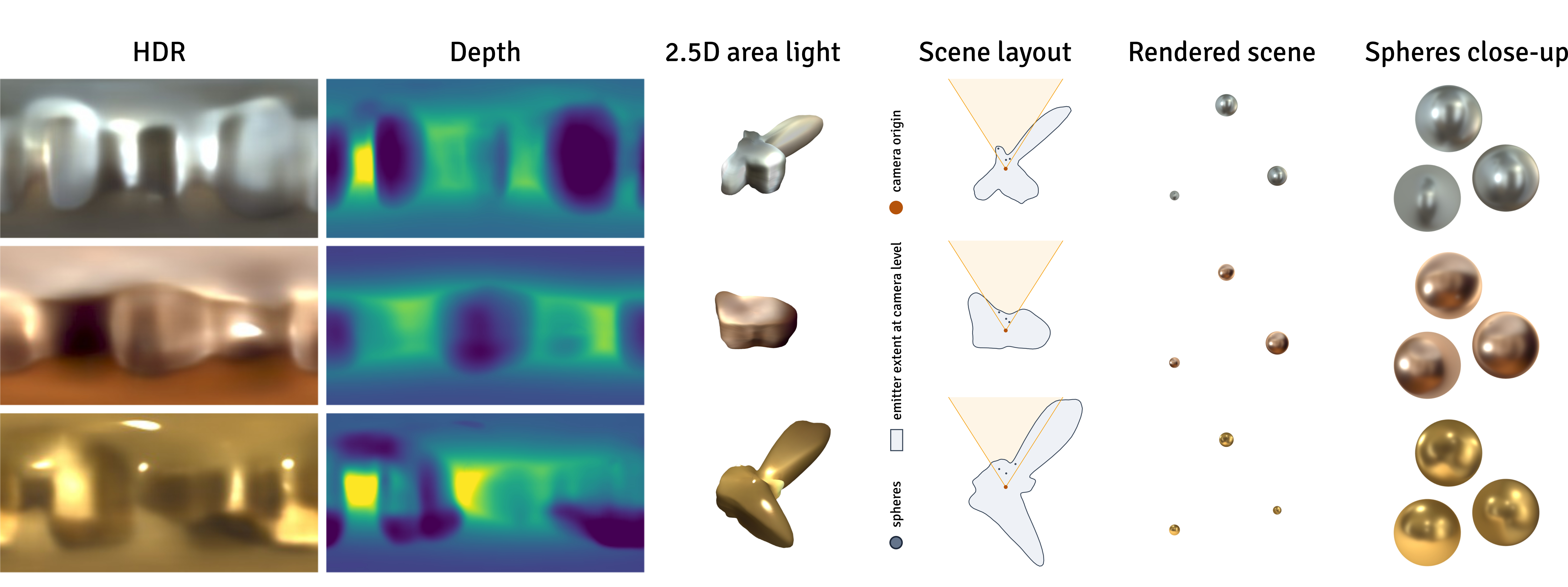}
    \vspace{-0.2cm}
    \caption{\label{fig:random_samples}
    Random samples from our indoor illumination prior. Decoded HDR radiance and depth maps form 2.5D area lights for illuminating a scene with glossy spheres, demonstrating position-dependent effects. Viewing angles: $45^\circ$ elevation (emitter mesh), $90^\circ$ (scene layout). \vspace{-1em}}
\end{figure*}

We propose an illumination prior for indoor scenes: a variational autoencoder over HDR equirectangular panoramas augmented with an aligned radial depth channel.
A single global latent code $z$ summarizes an entire environment.
Decoding it yields a radiance map and a depth map, which we lift into a 2.5D area light usable directly as an emitter in a differentiable rendering framework.
Illumination can then be recovered from images by optimizing $z$ through the renderer, so the prior restricts the solution to plausible indoor environments while the depth channel makes the recovered illumination spatially varying.
Fig.~\ref{fig:method} provides an overview.

\subsection{Emitter construction}
\label{sec:method:emitter}
Let $L \in \mathbb{R}^{H \times W \times 3}_+$ be an HDR panorama in linear radiance and $D \in \mathbb{R}^{H \times W}_+$ its aligned radial depth map, both given in the equirectangular domain.
We construct from the pair $(L, D)$ a 2.5D area light, represented as a closed triangle mesh with an emissive texture, by exploiting the spherical parametrization of the equirectangular image domain.

Each normalized pixel coordinate $(u, v) \in [0, 1]^2$ maps to an azimuth angle $\phi = 2 \pi u$ and a polar angle $\theta = \pi v$. These angles define a unit camera-ray direction vector,
\begin{equation}
    \omega (u,v) = (\sin\phi \sin\theta, \cos\theta, -\cos\phi \sin\theta).
\end{equation}
We lift the 2D image coordinates into 3D space by scaling the ray direction by its corresponding depth value to obtain the vertex position:
\begin{equation}
    p(u, v) = D(u,v)\, \omega(u,v).
\end{equation}
The mesh topology is constructed by triangulating the regular quad grid defined by neighboring pixels.
The image domain is sampled at pixel centers, so the first and last rows lie at $\theta = \tfrac{\pi}{2H}$ and $\theta = \pi - \tfrac{\pi}{2H}$ rather than at the poles themselves, which leaves two small open rings around the polar axis.
To seal these open boundaries and close the mesh topology, we introduce dedicated pole vertices computed as the centroids of their respective boundary rows:
\begin{equation}
\begin{aligned}
p_N &= \frac{1}{W}\sum_{i=0}^{W-1} p\left(\frac{i + 0.5}{W},\; \frac{0.5}{H}\right), \\
p_S &= \frac{1}{W}\sum_{i=0}^{W-1} p\left(\frac{i + 0.5}{W},\; 1 - \frac{0.5}{H}\right).
\end{aligned}
\end{equation}
These vertices are then connected to the top and bottom rows using triangle fans.
Since the mesh is derived from the equirectangular parameterization, the normalized image coordinates $(u,v)$ naturally serve as texture coordinates, indexing the HDR panorama $L$ used as the emissive texture.

The construction is a differentiable function of $(L, D)$ with fixed topology and fixed texture coordinates: $D$ moves the vertex positions and $L$ is the emissive texture, so gradients from a rendered image reach both.
At inference we apply it to the decoded pair $(\hat{L}, \hat{D})$ rather than to a captured one, which is what turns the latent into a renderable, spatially-varying light source.

\subsection{Panoramic convolutional VAE}
We model the joint distribution of radiance and depth with a convolutional VAE~\cite{kingma2014vae} that compresses a panorama into a single global latent $z \in \mathbb{R}^{512}$.
A global code enforces holistic consistency across the environment and gives downstream optimization a compact, low-dimensional search space.

The encoder is a convolutional stem followed by four stages of strided residual downsampling~\cite{he2016resnet} with GroupNorm~\cite{wu2018groupnorm} and SiLU~\cite{elfwing2018silu}. 
The resulting feature map is projected to the parameters of a diagonal Gaussian posterior $q_\phi(z \mid x) = \mathcal{N}\!\left(\mu(x), \operatorname{diag} \sigma^2(x)\right)$, from which $z$ is drawn using the reparametrization trick.
 
The decoder symmetrically upsamples and refines features via residual blocks to reconstruct the panorama.
To propagate global context across all scales, we modulate decoder feature maps via Adaptive Group Normalization~\cite{karras2019style,dhariwal2021}, which also improves gradient flow during downstream latent optimization.

Two design choices adapt the architecture to the spherical topology of the equirectangular domain.
First, we follow~\cite{DA360} in using standard convolution kernels and replacing zero padding with circular padding~\cite{zioulis2018omnidepth}: wrap-around in longitude, and across each pole by reflecting the boundary rows and rolling them by $W/2$, which is where the neighbourhood beyond the pole actually lies on the sphere. Second, convolutions are translation-equivariant, whereas equirectangular statistics are strongly latitude-dependent.
We therefore make latitude explicit by concatenating the cosine and sine of the polar angle, $(\cos\theta, \sin\theta)$, as two additional channels to the inputs of both the encoder and the decoder~\cite{liu2018coordconv}.

\subsection{Training objective}
\label{sec:method:loss}

To manage the wide dynamic range, we apply non-linear mappings followed by per-channel standardization with fixed dataset statistics, giving the four-channel network input $x$.
Radiance is transformed via $\log(1+L)$ to compress high values without over-allocating precision to negligible dark regions.
Depth spans a narrower range and never vanishes, so $\log D$ preserves uniform relative precision across all distances.
We write $\hat{x}$ for the decoder output and $(\hat{L}, \hat{D})$ for the linear-domain quantities defining an emitter in Sec.~\ref{sec:method:emitter}.

Unlike existing illumination priors~\cite{gardner2023reni++,walker2026veni}, which use scale-invariant objectives, we move the invariance out of the loss and into the data.
Every panorama is exposure calibrated to unit log-average luminance before training, so a radiance value means the same thing across the dataset.
This allows the terms below to be absolute, and in particular lets $\mathcal{L}_{\text{peak}}$ to constrain light-source intensity in physical units.
Since a single observation determines illumination only up to an overall factor, downstream applications optimize a per-image exposure on the decoded radiance jointly with the latent, as in RENI++ and VENI.
The full training objective is
\begin{equation}
\mathcal{L}_{\text{train}
} =
  \mathcal{L}_{1}^{L} + \mathcal{L}_{1}^{D}
  + 2\,\mathcal{L}_{\nabla}^{L} + \mathcal{L}_{\nabla}^{D} + 0.1\,\mathcal{L}_{\text{peak}}
  + \beta_t\, \mathcal{L}_{\text{KL}},
\label{eq:objective}
\end{equation}
where $\mathcal{L}_{1}^{L}$ and $\mathcal{L}_{1}^{D}$ are the pixel-wise $L_1$ errors on the radiance and depth channels, and the remaining terms are described next.
An $L_1$ objective alone is indifferent to structure, yielding blurred radiance and flat depth.
Following VENI~\cite{walker2026veni}, and in the spirit of gradient losses for depth prediction~\cite{li2018megadepth}, we add a term on normalized Scharr derivatives~\cite{scharr1997numerische} accumulated over an $S = 4$ octave pyramid
\begin{equation}
\mathcal{L}_{\nabla} = \frac{1}{2S} \sum_{o=1}^{S} \frac{1}{N_o} \sum_{i} \sum_{d \in \{u, v\}} \bigl| \nabla_d \hat{y}^{(o)}_i - \nabla_d y^{(o)}_i \bigr|,
\end{equation}
where $\nabla_u, \nabla_v$ are the horizontal and vertical Scharr operators, $i$ ranges over the $N_o$ pixels and channels of octave $o$, and $y$ is $x^{L}$ or $x^{D}$ for $\mathcal{L}_{\nabla}^{L}$ and $\mathcal{L}_{\nabla}^{D}$ respectively.

Downstream rendering is highly sensitive to peak illumination, which neither of the terms above constrains. We therefore introduce a loss focused on the high-intensity tail. Let $M$ denote a binary mask covering the union of the top $0.5\%$ brightest pixels (by luminance) in both $L$ and $\hat{L}$ for each sample. Penalizing both omitted and hallucinated emitters, the log-space $L_1$ loss over $M$ is defined as
\begin{equation}
\mathcal{L}_{\text{peak}} = \frac{1}{3|M|} \sum_{i \in M} \sum_{c \in \{r,g,b\}}
    \bigl| \log(\hat{L}_{i,c}) - \log(L_{i,c}) \bigr|.
\end{equation}

We regularize the latent space toward a standard normal prior $p(z) = \mathcal{N}(0, I)$ using the closed-form KL divergence:
\begin{align}
\mathcal{L}_{\text{KL}} &= D_{\mathrm{KL}}\!\left( q(z \mid x) \,\|\, \mathcal{N}(0, I) \right) \\
 &= \tfrac{1}{2} \sum_{j} \left( \mu_j^2 + \sigma_j^2 - 1 - \log \sigma_j^2 \right),
\end{align}
summed over all latent dimensions. To prevent posterior collapse, the loss weight is linearly annealed as $\beta_t = \beta \min(t / T_{\mathrm{w}}, 1)$, with warmup over $T_{\mathrm{w}} = 1{,}000$ steps and target weight $\beta = 10^{-3}$.

\subsection{Prior-guided illumination recovery}
\label{sec:method:optim}
We leverage the prior to recover illumination via analysis-by-synthesis: a hypothesized environment is decoded, rendered through a differentiable renderer \cite{Mitsuba3}, and compared against the observation.
To isolate illumination from other appearance factors, we consider a single low-dynamic-range (LDR) observation $I$ of objects with known geometry, material properties, and poses, captured under a calibrated camera, and address a more general setup in Sec. \ref{sec:exp:real}. 

With the pre-trained decoder fixed, optimization is restricted to the latent code $z$ and an unknown log-scale exposure parameter $s$, yielding a constrained parameter space of $512 + 1$ degrees of freedom.
Decoding $z$ and scaling the radiance by the exposure $e^s$ produces 
\begin{equation}
    \bigl(\hat{L}_s,\, \hat{D}\bigr) = \bigl(e^{s}\, \hat{L}(z),\; \hat{D}(z)\bigr),
\end{equation}
which defines the 2.5D area light emitter described in Sec.~\ref{sec:method:emitter}, and is used to render a view of the scene $R(z, s)$. The target observation is low dynamic range, so saturated pixels record only that the radiance exceeded the sensor's limit. We apply the same censoring to the render before computing the losses:
\begin{equation}
    \tilde{R}_i =
    \begin{cases}
        \min(R_i, 1),        & \max_c I_{i,c} \geq \tau, \\
        \min(R_i, c_{\max}), & \text{otherwise},
    \end{cases}
\end{equation}
so that a saturated pixel constrains the illumination only from below, while an unsaturated one is left intact up to a loose ceiling $c_{\max}$ that prevents a single over-bright sample from dominating the gradient.

Having $\Omega$ represent the set of pixels covered by the objects, we compute the photometric loss as
\begin{equation}
\begin{aligned}
    \mathcal{L}_{\text{img}} = &\; \frac{1}{|\Omega|} \sum_{i \in \Omega} \|\tilde{R}_i - I_i\|^2 \\
    &+ \frac{\lambda_{\log}}{|\Omega|} \sum_{i \in \Omega} \|\log(\tilde{R}_i + \epsilon) - \log(I_i + \epsilon)\|^2 .
\end{aligned}
\end{equation}

To prevent the 2.5D area light from colliding with scene geometry, we add a repulsion penalty:
\begin{equation}
\mathcal{L}_\mathrm{rep} = \frac{1}{\vert{}Q\vert{}} \sum_{q \in Q} \max(m - g(q),\, 0)
\end{equation}
where $g(q)$ is the signed clearance of an object point $q$ to the emitter (positive inside, negative outside) and $m$ is the safety margin. The full objective minimized over $(z, s)$ is
\begin{equation}
    \mathcal{L}_{\text{optim}} = \mathcal{L}_{\text{img}} + \lambda_{\text{rep}}\, \mathcal{L}_{\text{rep}}
\end{equation}

\section{Experiments}

This section presents the training, evaluation protocol, and empirical performance of our proposed prior across various settings. First, Sec.~\ref{sec:exp:prior} assesses the prior in isolation as a joint generative model of radiance and depth. Next, Sec.~\ref{sec:exp:synth} evaluates its integration into the inverse-rendering optimization loop (Sec.~\ref{sec:method:optim}) on synthetic scenes. 
Finally, Sec.~\ref{sec:exp:real} tests performance on real-world photographs with a more challenging setup. 
Further results, including an ablation study, are presented in the supplementary material, Sec. ~\ref{sup:results}.

\begin{figure}
    \centering
    \includegraphics[width=0.99\linewidth]{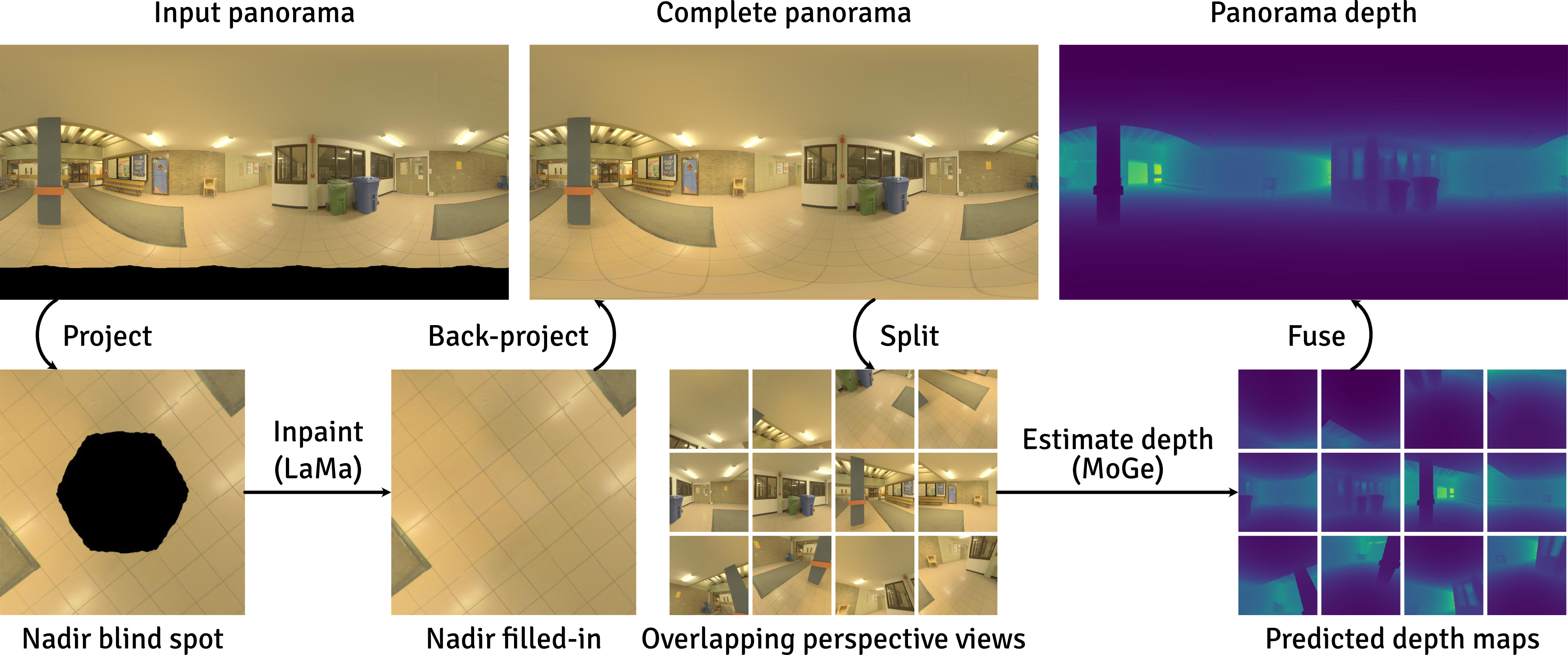}
    \caption{\label{fig:laval_dataset}
    Preprocessing of the Laval Indoor HDR dataset \cite{gardner2017indoor} used to train our illumination prior. We inpaint \cite{suvorov2022resolution} the blind spot of the camera and estimate a panoramic depth map \cite{wang2025moge2}.
    \vspace{-1.5em}}
\end{figure}

\vspace{-1em}
\paragraph{Training data.}
Our illumination prior requires indoor HDR panoramas paired with depth maps, but no real-world dataset provides such pairs at sufficient scale.
We therefore augment the Laval Indoor HDR dataset~\cite{gardner2017indoor} with pseudo-ground-truth depth from an off-the-shelf monocular estimator~\cite{wang2025moge2}, and inpaint the nadir blind spot of the capture setup~\cite{suvorov2022resolution} (see Fig.~\ref{fig:laval_dataset}).
This yields $2233$ training samples.

\vspace{-1em}
\paragraph{Baselines.} We compare against three types of methods:

\textit{Single-image illumination estimators.} These methods capture spatially varying indoor illumination, but their output is a direct estimate rather than a parameterization, meaning there is no low-dimensional latent code that can be optimized against the observation. We consider DepthLight~\cite{manus2025depthlight}, which uses a representation similar to our 2.5D area light but is obtained via outpainting, and 4D Lighting~\cite{tong2025spatio4dsig}, which inpaints chrome probes into the input image. 

\textit{Direct optimization (Smoothness).} To isolate the prior's contribution from the representation and fitting procedure, we directly optimize radiance and depth maps under the same setup ($131{,}072$ parameters vs. $513$ for our prior at the same map resolution). To test the role of depth, we replace the mesh with a fixed-radius spherical proxy or remove it entirely via an environment map. Because unregularized optimization yields very noisy estimates, we apply NeRFactor's smoothness constraint~\cite{zhang2021nerfactor} on log-radiance and log-depth.

\begin{table}
\centering
\begin{adjustbox}{width=\columnwidth}
\small
\setlength{\tabcolsep}{3.5pt}
\begin{tabular}{c l ccccc}
\toprule
& \multirow{2}{*}{Method} & \textbf{PSNR} $\uparrow$ & \textbf{SSIM} $\uparrow$ & \textbf{LPIPS} $\downarrow$ & \textbf{PSNR} $\uparrow$ & \textbf{AbsRel} $\downarrow$ \\
&  & \multicolumn{3}{c}{----------- \: LDR \: -----------}  &  HDR & Depth \\
\midrule
\multirow{3}{*}{\rotatebox[origin=c]{90}{Encoder}} 
&VENI~\cite{walker2026veni}      & 13.84 & 0.554 & 0.724 & 31.76 & -- \\
&VENI\textcolor{mygreen}{+D (ours)}                  & 12.60 & 0.532 & 0.745 & 31.51 & 0.098 \\
&\textcolor{mygreen}{Ours}       & \colorbox{tabfirst}{19.14} & \colorbox{tabfirst}{0.661} & \colorbox{tabfirst}{0.575} & \colorbox{tabfirst}{33.90} & \colorbox{tabfirst}{0.049} \\
\midrule
\multirow{3}{*}{\rotatebox[origin=c]{90}{Latent fit}}
&VENI~\cite{walker2026veni}      & 21.37 & 0.701 & 0.526 & 33.16 & -- \\
&VENI\textcolor{mygreen}{+D (ours)}                 & 21.92 & 0.713 & 0.520 & 33.64 & 0.092 \\
&\textcolor{mygreen}{Ours}       & \colorbox{tabfirst}{25.34} & \colorbox{tabfirst}{0.751} & \colorbox{tabfirst}{0.422} & \colorbox{tabfirst}{37.12} & \colorbox{tabfirst}{0.045} \\
\midrule
\multirow{3}{*}{\rotatebox[origin=c]{90}{\scriptsize{Pixel-level}}} & Smooth. HDR & 	\colorbox{tabfirst}{25.43} & \colorbox{tabfirst}{0.871} & 0.255 & \colorbox{tabfirst}{35.18} & -- \\
& Smooth. HDRD & 24.47 & 0.849	& 0.296 & 34.65 & \colorbox{tabfirst}{0.018} \\
& DPI~\cite{lyu2023diffpostillum} & 20.94 & 0.844 & \colorbox{tabfirst}{0.183} & 33.63 & -- \\
\bottomrule
\end{tabular}
\end{adjustbox}
\caption{Panorama reconstruction on the evaluation set. Encoder: a single forward pass through the VAE. Latent fit: the latent is optimized against the target with the decoder frozen. Pixel-level: reference points with no bottleneck, their scores report the regularization they impose (Smoothness) or the strength of the guidance signal (DPI), not the fidelity of a compact representation.
\vspace{-1em}}
\label{tab:prior_recon}
\end{table}

\begin{figure*}
    \centering
\includegraphics[width=0.99\linewidth]{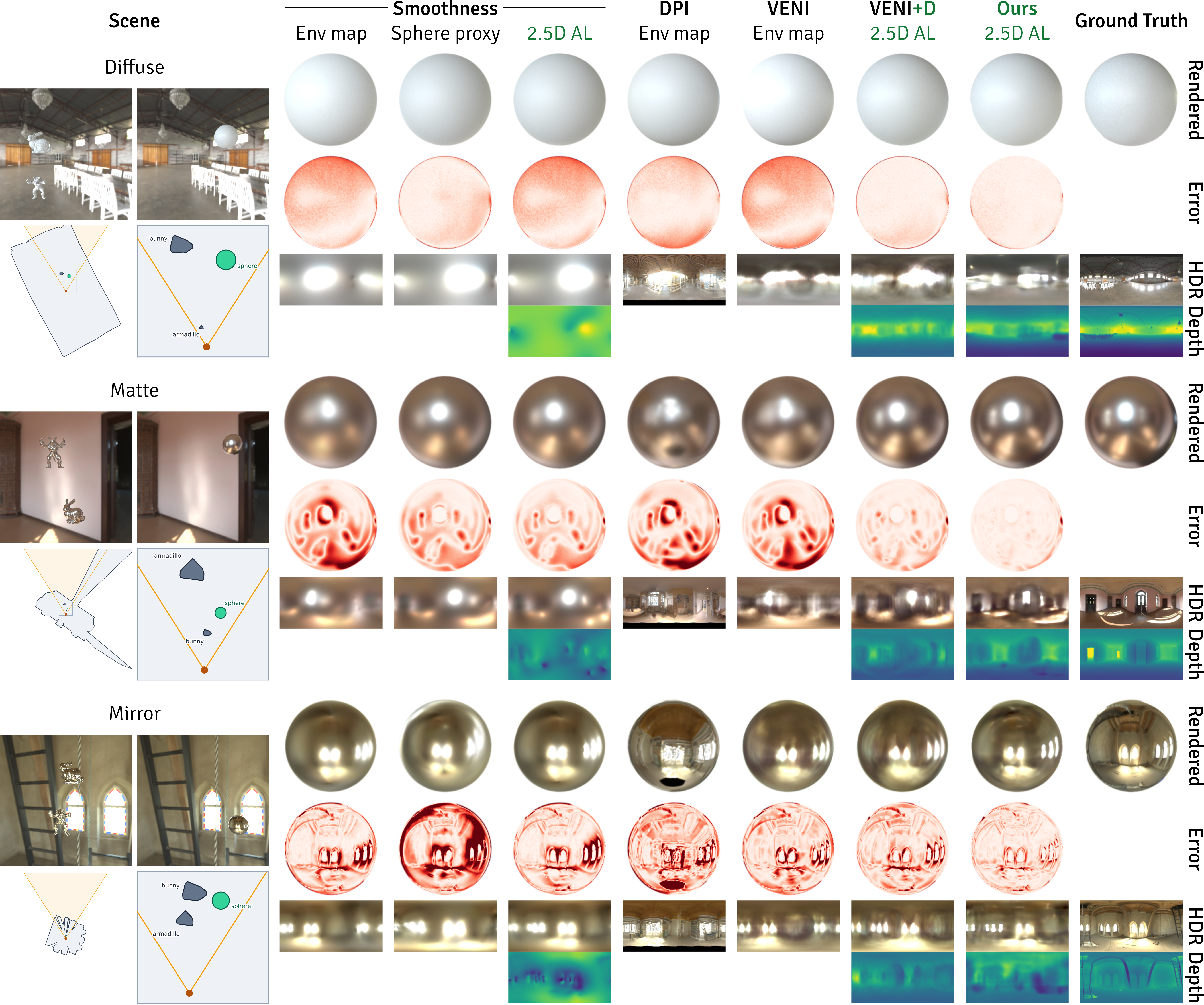}
    \caption{\label{fig:results_synthetic}
    Qualitative comparison across three scenes (one per material). Left: Input view used for fitting, held-out evaluation view with the probe sphere, and top-down room layouts (full and close-up) showing camera frustums and object placements. Right (per method): Re-rendered probe, its error against the ground truth on the tone-mapped view, recovered panorama and depth where applicable.} 
    \vspace{-1em}
\end{figure*}

\begin{table*}
\centering
\begin{adjustbox}{width=0.95\linewidth}
\small
\begin{tabular}{cc ccc ccc ccc}
\toprule 
\multirow{2}{*}{Method} &
\multirow{2}{*}{Representation}
& \multicolumn{3}{c}{\textbf{PSNR LDR} $\uparrow$} & \multicolumn{3}{c}{\textbf{RMSE HDR} $\downarrow$} & \multicolumn{3}{c}{\textbf{RGB Angular Error} $\downarrow$} \\
\cmidrule{3-5} \cmidrule{6-8} \cmidrule{9-11} 
 &  & Diffuse & Matte & Mirror & Diffuse & Matte & Mirror & Diffuse & Matte & Mirror \\
\midrule

 & Env map
& 26.15 & 20.63 & 17.31 & 0.148 & 0.650 & 2.794 & 1.510 & 3.000 & 4.560\\

 Smoothness & Sphere proxy
 & 29.63 & 23.37 & 19.04 & 0.122 & 0.555 & 2.613 & 1.251 & 2.573 & \colorbox{tabsecond}{3.974}\\

  & \textcolor{mygreen}{2.5D AL}
 & 28.97 & 24.71 & 19.77 & 0.132 & 0.530 & 2.589 & \colorbox{tabsecond}{1.112} & \colorbox{tabsecond}{2.367} & \colorbox{tabfirst}{3.767} \\

DPI~\cite{lyu2023diffpostillum} & Env map
& 24.25 & 19.51 & 15.61 & 0.176 & 0.751 & 2.845 & 1.818 & 4.057 & 5.758\\

VENI~\cite{walker2026veni} & Env map
& 26.07 & 20.66 & 17.32 & 0.144 & 0.621 & 2.766 & 1.599 & 3.290 & 5.289\\

VENI\textcolor{mygreen}{+D (ours)} & \textcolor{mygreen}{2.5D AL}
& \colorbox{tabsecond}{32.75} & \colorbox{tabsecond}{28.24} & \colorbox{tabsecond}{21.17} & \colorbox{tabsecond}{0.113} & \colorbox{tabsecond}{0.391} & \colorbox{tabsecond}{2.364} & \colorbox{tabfirst}{1.076} & 2.778 & 4.813\\

\textcolor{mygreen}{Ours} & \textcolor{mygreen}{2.5D AL}
& \colorbox{tabfirst}{33.63} & \colorbox{tabfirst}{30.41} & \colorbox{tabfirst}{22.85} & \colorbox{tabfirst}{0.089} & \colorbox{tabfirst}{0.272} & \colorbox{tabfirst}{2.150} & 1.176 & \colorbox{tabfirst}{2.269} & 4.372\\

\bottomrule
\end{tabular}
\end{adjustbox}
\caption{Quantitative evaluation on probe spheres. Results are averaged across three materials over 25 synthetic scenes. The 2.5D area light representation performs best by modeling spatially varying illumination. \colorbox{tabfirst}{Best} and \colorbox{tabsecond}{second best} results are highlighted.}
\vspace{-0.5em}
\label{tab:reconstruction}
\end{table*}

\begin{figure}
    \centering
    \includegraphics[width=0.99\linewidth]{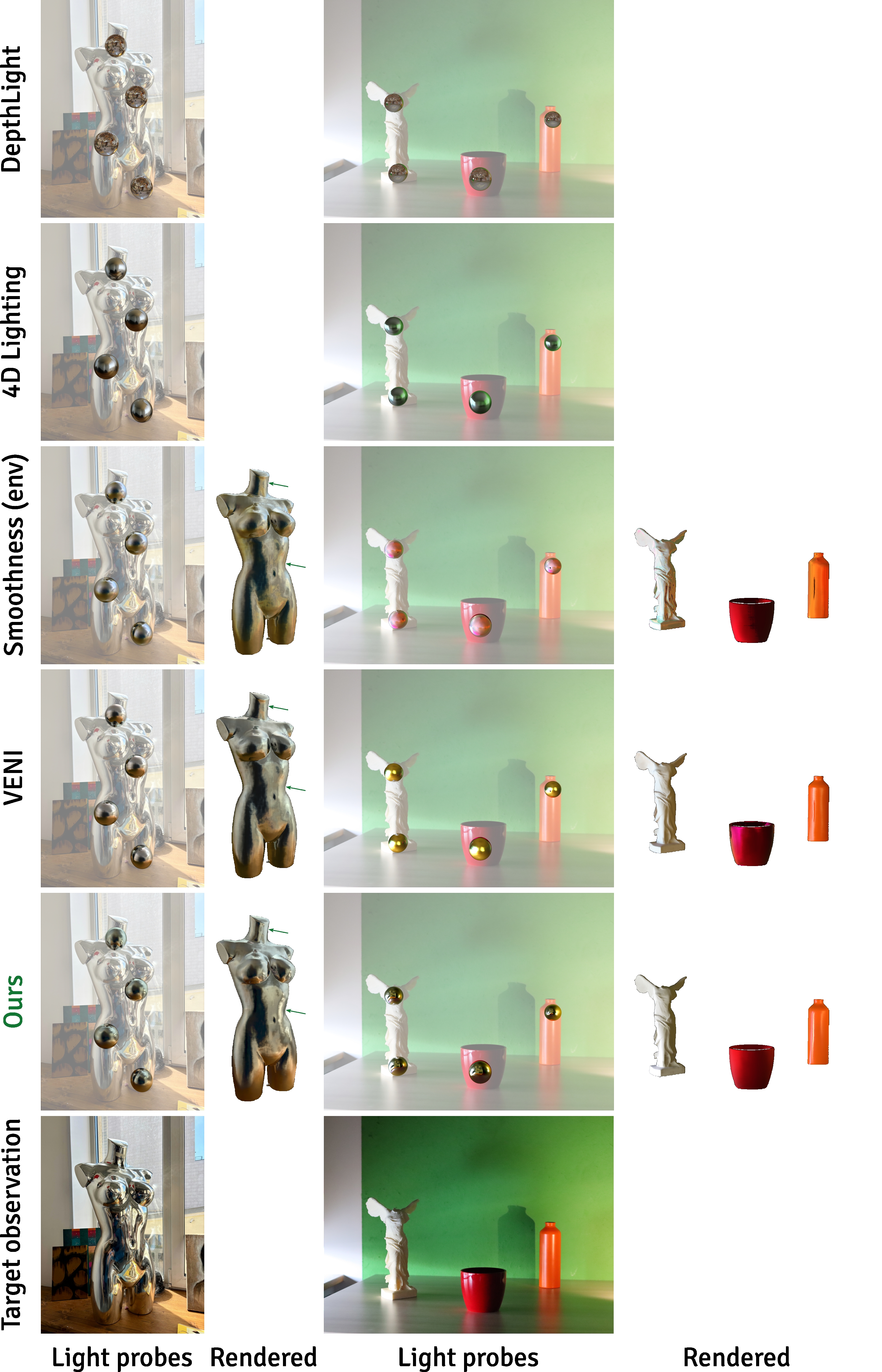}
    \caption{\label{fig:real}
    Illumination recovery on real photographs. For each method we insert virtual mirror probes at several positions in the reconstructed scene and render them into the photograph, and we re-render the segmented objects with the jointly optimized materials under the recovered illumination. DepthLight~\cite{manus2025depthlight} and 4D Lighting~\cite{tong2025spatio4dsig} predict illumination directly and do not estimate materials, hence no re-rendering. Smoothness and VENI~\cite{walker2026veni} parameterize lighting as an environment map, so every probe sees the same radiance regardless of where it stands, whereas our optimized 2.5D area light yields position-dependent probe appearance.
    \vspace{-1em}}
\end{figure}

\textit{Illumination priors, fitted through the same renderer.} 
DPI~\cite{lyu2023diffpostillum}, using their original implementation, weights and fitting procedure while bypassing the material optimization.
VENI~\cite{walker2026veni}, trained on our indoor split using the original implementation with matching latent size and resolution. VENI+D, no spatially-aware illumination prior exists to compare against, so we build one. We extend VENI with a depth channel, so that its latent decodes into a radiance–depth pair that can be lifted into a 2.5D area light. This isolates the contribution of our architecture from that of the representation: VENI+D shares our emitter and our fitting procedure, and differs only in the prior, making it a representative ablation. 

\vspace{-1em}
\paragraph{} Additional details about the experimental setup are included in Sec. \ref{sup:exp_details} of the supplementary material.

\subsection{Illumination prior evaluation}
\label{sec:exp:prior}
We first evaluate the prior directly, without a renderer.
The evaluation set consists of $25$ real indoor HDR panoramas from a different source~\cite{polyhaven}, with pseudo-ground-truth depth from the same pipeline as the training data~\cite{wang2025moge2}.
Following the VENI~\cite{walker2026veni} protocol, we report PSNR, SSIM, and LPIPS in LDR tone-mapped space, as well as PSNR in linear HDR space.
We also include the Absolute Relative Error for measuring the depth performance. 
Tab.~\ref{tab:prior_recon} shows that our model outperforms VENI in both the autoencoder setting and decoder-only latent optimization.
Smoothness scores well here by construction: with one parameter per target value, its metrics report the level of smoothing imposed rather than representation quality, and are readable only alongside Sec.~\ref{sec:exp:synth}.
Similarly, DPI~\cite{lyu2023diffpostillum} reconstructs through no bottleneck at all, as the target panorama is its measurement and guides the sampler at every step.

Fig.~\ref{fig:random_samples} shows draws from $\mathcal{N}(0, I)$, each decoded into radiance and depth maps, lifted to 2.5D area lights (Sec.~\ref{sec:method:emitter}), and used to illuminate glossy spheres. The generated samples depict coherent rooms with diverse layouts and distinct light sources, while the emitter's geometry ensures the spheres' appearance varies naturally based on their 3D position.

\subsection{Inverse rendering on synthetic scenes}
\label{sec:exp:synth}
\noindent\textbf{Dataset.} We build a dedicated dataset based on the $25$ indoor panoramas~\cite{polyhaven} with estimated depth.
Unlike in the previous evaluation, we use a higher resolution of $512 \times 1024$ and construct 2.5D area emitters as in Sec.~\ref{sec:method:emitter}.
With the rendering camera placed at the emitter origin, we randomly position three objects at non-overlapping image coordinates, with their depths stratified across the range defined by the emitter extent.
The bunny and armadillo form the target observation used during optimization (Sec.~\ref{sec:method:optim}), and the evaluation is performed on a view containing a sphere.
Every scene is rendered in three materials spanning the reflective spectrum: gray-diffuse, silver-matte, and silver-mirror, amounting to $75$ cases ($25$ scenes $\times$ $3$ materials).

\noindent\textbf{Metrics.} The optimized illumination representation is used as the emitter for rendering the evaluation probe on which we compute PSNR in tone-mapped space, RMSE in linear space and mean per-pixel RGB angular error in degrees~\cite{legendre2019deeplight}.  RMSE measures intensity errors by heavily penalizing brightness discrepancies in light sources, while angular error isolates color accuracy by weighting all pixels equally regardless of scale. 

Quantitative results in Tab.~\ref{tab:reconstruction} demonstrate that near-field modeling yields substantial gains, as evidenced by the performance of both the smoothness baseline and VENI across different emitter representations. 
Our prior leads on almost every case and metric, extending the advantage it showed
over VENI+D in the direct evaluation (Tab.~\ref{tab:prior_recon}) to the
inverse-rendering setting, albeit by a narrower margin.
The only metric on which our model does not lead is the angular error, for the diffuse and mirror probes. As this metric is scale-free, it can be won by an environment whose radiance is the right color but the wrong strength, which shows in the RMSE and PSNR metrics. The qualitative results in Fig. \ref{fig:results_synthetic} further confirm the advantage of the data-driven priors over simple smoothness constraints. In the diffuse case, the directly optimized radiance is reduced to just a few bright blobs, while our prior produces more realistic details. The depth channel also highlights that the smoothness prior degenerates into a low-frequency field carrying no room structure, whereas our decoded depth resembles the reference layout.  

A fair comparison against DepthLight~\cite{manus2025depthlight} and 4DLighting~\cite{tong2025spatio4dsig} is not possible in this setting: both require a representative background, which we mask out during optimization, and the optimization-based arms would in any case hold an unfair advantage, isolating illumination with ground-truth object geometry and materials.
For completeness we report their results on this dataset in the supplementary material.

\subsection{Application on real images}
\label{sec:exp:real}

The synthetic evaluation isolates illumination by assuming known geometry, materials and poses.
Single-view applications afford none of these.
Geometry can be recovered through generative reconstruction \cite{Ardelean2025Gen3DSR, chen2026sam, yin20263d} or partially from monocular depth \cite{wang2025materialist, careaga2025physically}, but materials must be estimated jointly with lighting, reintroducing the ambiguity that priors are meant to resolve.
We therefore evaluate our prior in a simplified analysis-by-synthesis setting on real images.

\vspace{-1em}
\paragraph{Setup.} From a single image we obtain camera intrinsics and a depth-based scene mesh \cite{wang2025moge2}, and segment the objects of interest. Each object is assigned a single spatially-uniform material under an albedo–roughness–metallic parameterization. We then minimize the photometric loss over the material parameters, the exposure scale $s$, and the latent code $z$ of our prior, which decodes into the 2.5D area light illuminating the reconstructed geometry. 

Real photographs offer no ground-truth illumination, so in
Fig.~\ref{fig:real} we assess the recovered lighting through its
consequences: virtual mirror probes inserted at several positions in
the reconstructed scene, and, where applicable, the objects re-rendered
with jointly optimized materials. Smoothness and
VENI~\cite{walker2026veni} parameterize lighting as an environment map,
so every probe returns the same radiance regardless of position,
which is visibly inconsistent with the surrounding photograph. Our 2.5D
area light instead reflects the part of the room each probe actually
faces, matching the scene layout.
DepthLight~\cite{manus2025depthlight} and 4D
Lighting~\cite{tong2025spatio4dsig} also yield position-dependent
probes, but their illumination is produced from generative image
priors rather than by rendering a hypothesis against the
observation, so it is never constrained by the appearance of the
objects themselves and cannot be recovered jointly with their
materials. While this is a simplified real-world setting, it shows that
our prior transfers beyond the controlled synthetic setup and remains
usable when geometry and materials are themselves estimated.

\section{Discussion}

\paragraph{Limitations.}
A 2.5D area light carries one radiance and one distance per direction, so depth discontinuities are bridged by triangles that weld foreground to background. Moreover, emitters standing among the objects rather than in the enclosure cannot be properly expressed.
The equirectangular parametrization adds to this by allocating many small triangles near the poles and few near the equator, where most details lie.
A further limitation lies in the fitting, where we rely on a repulsion term to keep the emitter from intersecting the scene content.
This works well in most cases, but it requires a set of scene points against which the clearance is measured, and it addresses a symptom rather than a cause: a perfectly recovered environment would enclose the scene without intersecting it, so no such term would be needed.

\vspace{-1em}
\paragraph{Conclusion.}
We introduced a learned illumination prior that represents indoor lighting as a 2.5D
area light rather than a far-field environment map. A single latent code decodes into
paired HDR radiance and radial depth, lifted into an emissive mesh directly usable with a
differentiable renderer. The finite extent of the emitter makes the
resulting light field vary with position, while the learned decoder keeps latent
optimization on the manifold of plausible rooms instead of the noisy solutions that
unconstrained pixel-level fitting produces. We show that this recovers illumination more faithfully than environment-map priors fitted through the same renderer across diverse indoor scenes and 
materials.

\vspace{0.5\baselineskip}
{
\textbf{Acknowledgement.}
We thank Paul Himmler and \mbox{Timotei} Ardelean for their constructive feedback during project development and manuscript preparation. 
The authors gratefully acknowledge the scientific support and HPC resources provided by the Erlangen National High Performance Computing Center (NHR@FAU) of the Friedrich-Alexander-Universität Erlangen-Nürnberg (FAU) under the NHR project b315dc RoomLight. NHR funding is provided by federal and Bavarian state authorities.
}

{
    \small
    \bibliographystyle{ieeenat_fullname}
    \bibliography{bibliography}
}

\clearpage
\appendix

\twocolumn[%
  \centering
  {\Large\bfseries RoomLight: A 2.5D Illumination Prior for Indoor Environments\par}
  \bigskip
  {\large Supplementary Material\par}
  \bigskip\bigskip
]

\noindent

\section{Representation motivation}
\label{sup:motivation}

\begin{figure*}
    \centering
\includegraphics[width=0.99\linewidth]{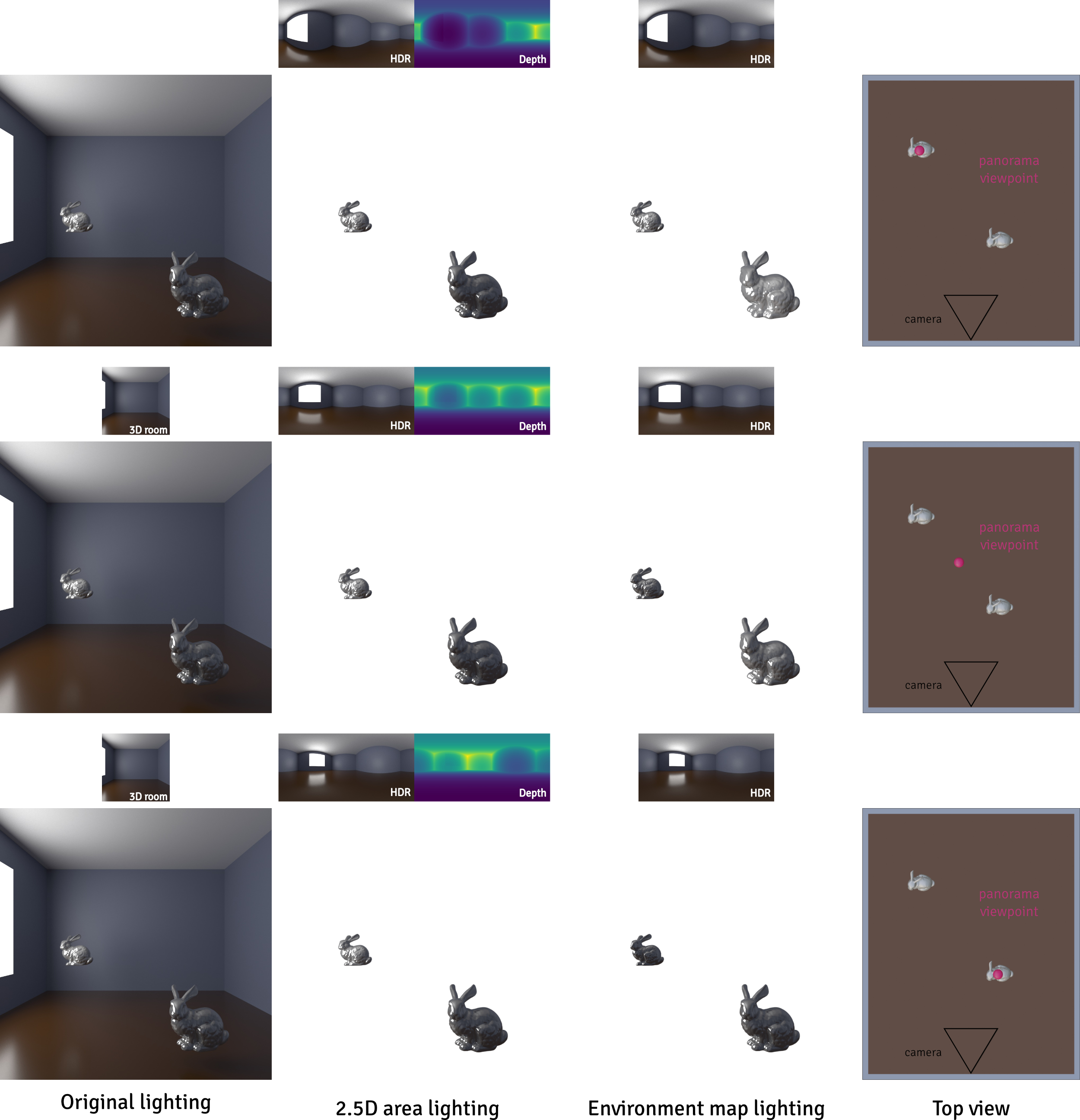}
    \caption{\label{fig:motivation_extended}
    Comparison of object relightings in a simple indoor environment, using a different panorama viewpoint for each row. Conventional environment map lighting assumes distant illumination and cannot simultaneously reproduce the appearance of both objects. In contrast, the 2.5D area lighting representation models spatially-varying illumination and closely reproduces the original rendering, regardless of the panorama viewpoint.} 
    \vspace{-1em}
\end{figure*}

Figure~\ref{fig:motivation_extended} extends Fig.~\ref{fig:motivation_varying} with panoramas captured at three locations: the center of each object and the midpoint between them. An environment map relights only the object at its capture location correctly, whereas the 2.5D area light reproduces both objects from every capture point.

\begin{figure}
    \centering
    \includegraphics[width=0.99\linewidth]{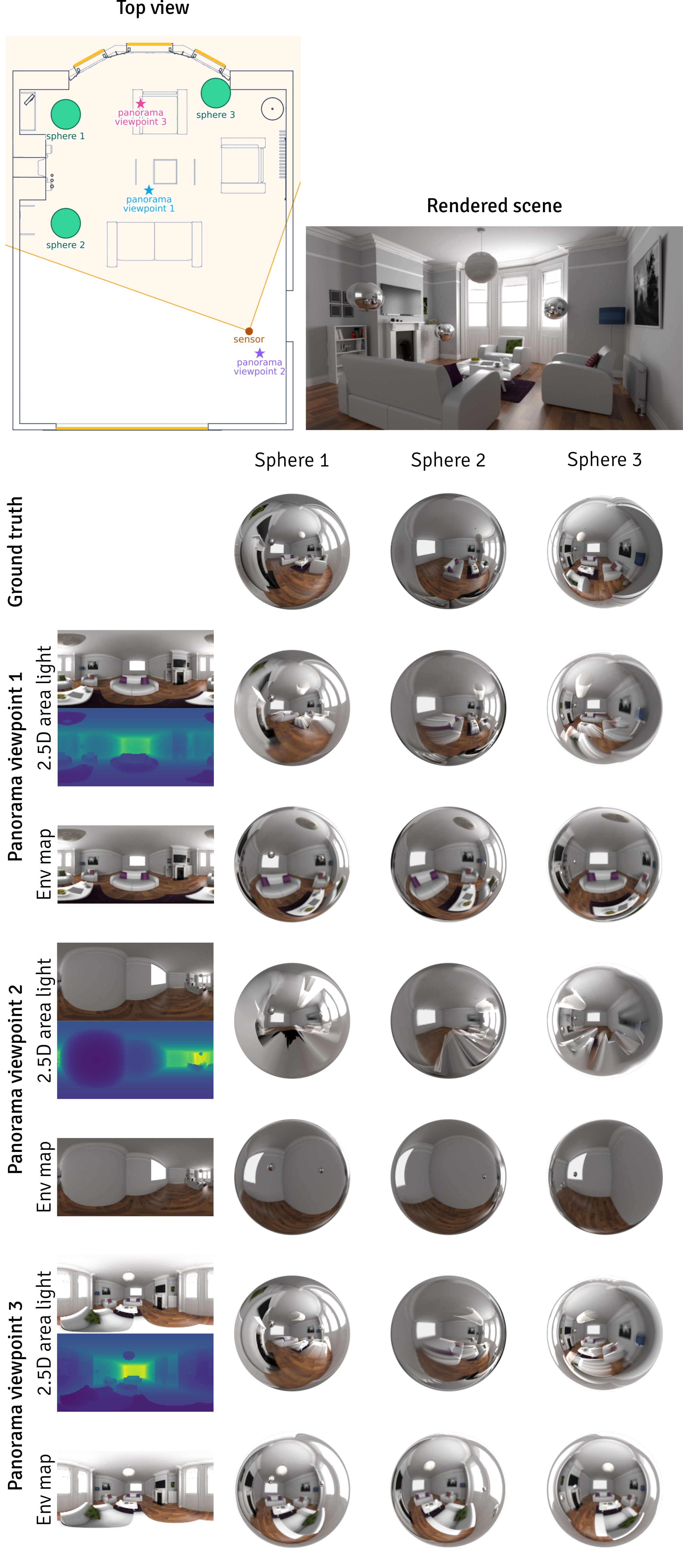}
    \caption{Comparing full light transport simulation in a 3D scene with 2.5D area lights and environment maps. The panoramas are captured from three different viewpoints (rows) and the appearance is analyzed across three mirror spheres (columns). The top view shows the location of the spheres and of the panorama viewpoints within the scene layout.}
    \label{fig:representation}
\end{figure}

\vspace{-1em}
\paragraph{Capacity of a single 2.5D area light.}
A 2.5D area light is only an approximation of the true illumination, as it captures a single depth per direction from the capture point. Surfaces hidden from that viewpoint are therefore absent, and depth discontinuities are bridged by triangles that are not a valid emitting surface.
To quantify how good the approximation is, we place three probe spheres in a path-traced living room, compute their appearance with the full light transport simulation, and compare it with relightings by 2.5D area lights and environment maps, as illustrated in Fig. \ref{fig:representation}. The emitters are built from the true radiance and depth captured at three viewpoints in the scene, and Table~\ref{tab:shell_capacity} aggregates their results.
The 2.5D area light achieves $2\times$ to $5\times$ lower RMSE than the environment map for every material.
Still, the error is much higher than that between re-renderings of the scene with different seeds (noise floor), which is the gap a 2.5D area light cannot represent. This bounds what any method built on this representation, ours included, can recover.

\begin{table}[h]
\centering
\begin{adjustbox}{width=0.99\linewidth}
\small
\begin{tabular}{l ccc}
\toprule
Emitter & Diffuse & Matte silver & Mirror \\
\midrule
Noise floor & 0.009 & 0.024 & 0.019 \\
\midrule
2.5D area light, $128 \times 256$ & 0.052--0.070 & 0.164--0.248 & 0.302--0.440 \\
Environment map & 0.130--0.377 & 0.864--1.493 & 1.427--2.455 \\
\bottomrule
\end{tabular}
\end{adjustbox}
\caption{Illumination error (linear radiance RMSE) for probe spheres lit by a single emitter versus the full-scene ground truth. The intervals are min--max across the three capture locations, averaged for the three probes. The first row establishes renderer noise as the difference between two renderings with different seeds.}
\vspace{-1em}
\label{tab:shell_capacity}
\end{table}

\section{Experimental setup}
\label{sup:exp_details}

\subsection{Training data preprocessing}
Additional details on the pipeline of Fig.~\ref{fig:laval_dataset}.

\vspace{-1em}
\paragraph{Nadir inpainting.} Panoramas captured with a tripod-mounted rig contain a nadir blind spot at the downward pole. 
To inpaint this region, we tone-map the HDR panorama to LDR using the extended Reinhard operator~\cite{reinhard2002}, then obtain a $512 \times 512$ top-down perspective view of the nadir area using gnomonic projection with an $80^\circ$ field of view. 
We inpaint the resulting hole with LaMa~\cite{suvorov2022resolution} and back-project the completed region into the panorama through the inverse gnomonic mapping. 
Finally, the Reinhard mapping is inverted to restore HDR values in the completed region, leaving the rest of the panorama unchanged.

\begin{figure*}
    \centering
\includegraphics[width=0.99\linewidth]{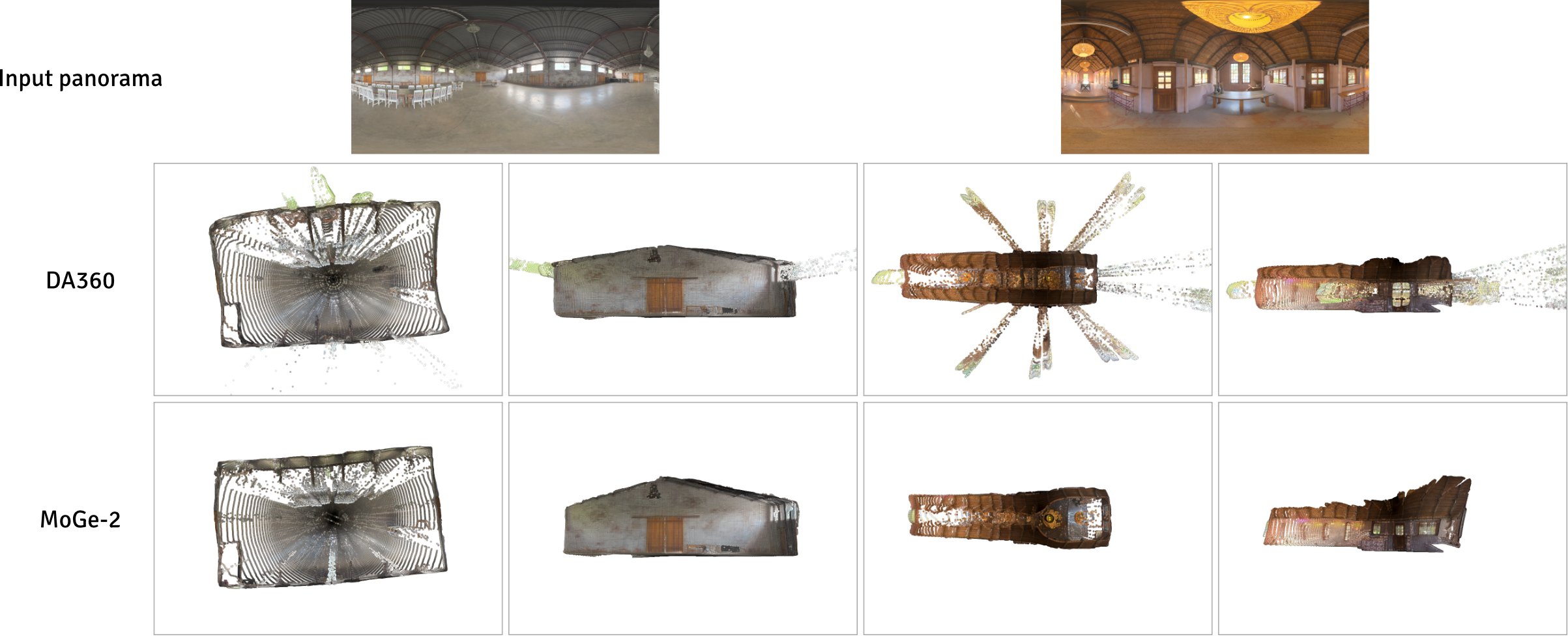}
    \caption{\label{fig:depth_comparison}
Point clouds unprojected from equirectangular depth predictions of
DA360~\cite{DA360} and MoGe-2~\cite{wang2025moge2}, shown in top and side orthographic views for two PolyHaven~\cite{polyhaven} scenes. Colors of the points are taken from the input panorama. MoGe-2 produces compact, closed room layouts, whereas DA360 places window and opening pixels at extreme depths, with points far beyond the room boundary.} 
    \vspace{-1em}
\end{figure*}

\vspace{-1em}
\paragraph{Panorama depth estimation.} While the domain-specific model DA360~\cite{DA360} estimates $360^\circ$ depth, it produces inconsistent predictions in transparent regions such as windows, with depth values exploding to high values (Fig.~\ref{fig:depth_comparison}). Instead, we adopt MoGe-2~\cite{wang2025moge2}, a depth estimator designed for perspective imagery. Following the authors' pipeline, we extract 12 overlapping tangent-plane perspective views from the tone-mapped panorama, infer depth on each view, project the predictions back to the equirectangular domain, and fuse them via global consistency optimization. Additionally, we enforce spherical pole constraints during optimization to preserve topological continuity near the poles.

\subsection{Synthetic scene design}
\label{sup:synth_design}

\paragraph{Emitter.} Each of the $25$ evaluation panoramas \cite{polyhaven} and its estimated depth are resampled to $512 \times 1024$ and lifted into a 2.5D area light (Sec.~\ref{sec:method:emitter}). This emitter lights both the fitted and the evaluation view, and every method is scored against renders produced with it.

\vspace{-1em}
\paragraph{Camera and objects.} A perspective camera ($65^\circ$ FoV, $1000 \times 1000$ resolution) pointing along $-z$ is placed in the origin of the 2.5D area light (the capture point of the panorama). The three objects are scaled to fit non-overlapping screen-space boxes (sizes $0.25$--$0.35$ frame width) and placed stratified ($[0.05, 0.25]$, $[0.4, 0.6]$, $[0.75, 0.95]$) within the depth range to the emitter mesh. The bunny and armadillo models of the target observation were sourced from the Stanford 3D Scanning Repository \cite{turk1994zippered, krishnamurthy1996fitting}.

\vspace{-1em}
\paragraph{Rendering.} Scenes are rendered in Mitsuba~3~\cite{Mitsuba3} with prb integrator at a maximum path depth of $4$. For Tab.~\ref{tab:reconstruction}, the ground-truth probe and the probe under each recovered illumination are rendered in linear HDR at $4096$ samples per pixel; the LDR PSNR is computed on the same renders after gamma-$2.2$ mapping and clipping to $[0, 1]$, over the probe pixels only.

\vspace{-1em}
\paragraph{Setup motivation.}
We render the ground truth with 2.5D area lights instead of fully modeled 3D rooms for three reasons.
First, the ground truth is real illumination: no collection of 3D indoor scenes with measured emitters exists at comparable diversity, and evaluating a prior trained on real panoramas on synthetic scenes would confound its quality with the domain gap.
Second, the problem is well-posed: an exact solution exists, so the scores measure the distance to it, and the recovered radiance and depth can be compared to the reference directly (Fig.~\ref{fig:results_synthetic}). Against a 3D room every 2.5D emitter is bounded by the ceiling of Tab.~\ref{tab:shell_capacity}, under which the differences between priors would be harder to read.
Third, the setup extends the protocol of prior work~\cite{gardner2023reni++, walker2026veni, lyu2023diffpostillum}, which evaluates on objects lit by a panorama, by giving the panorama a finite distance.

\subsection{Training details}
\label{sup:training}
The encoder is a $3 \times 3$ convolutional stem of width $64$ followed by four downsampling stages with channel multipliers $(1, 2, 4, 4)$, one residual block per stage, GroupNorm with $8$ groups and SiLU activations. The resulting $8 \times 16$ feature map is projected to $32$ channels and flattened into a linear layer emitting the $512$-dimensional posterior mean and log-variance. The decoder mirrors this structure, with the latent reshaped into the $8 \times 16$ starting map and injected at every normalization layer through AdaGN. 

Training runs for $300$ epochs on the $2233$ panoramas at $128 \times 256$ with a batch size of $32$, using AdamW~\cite{loshchilov2019adamw} with $(\beta_1, \beta_2) = (0.9, 0.99)$ and weight decay $0.01$. The learning rate is warmed up linearly to $2 \times 10^{-4}$ over $250$ iterations and decayed by a cosine schedule to $10^{-5}$; gradients are clipped to a global norm of $10$. Augmentations are a uniform azimuthal roll, a horizontal mirror, and a camera tilt of up to $\pm 7^\circ$, all of which preserve the equirectangular parametrization.

\subsection{Fitting details}
\label{sup:fitting}

\paragraph{Direct evaluation.}
For the latent-fit rows of Tab.~\ref{tab:prior_recon} we follow VENI's protocol~\cite{walker2026veni} for all models and optimize the 0-initialized latent using Adam optimizer to minimize the the squared error over the four normalized channels. The learning rate is decayed exponentially from $0.1$ to $0.01$ over $1500$ iterations (instead of $500$ as the fit keeps improving). Since depth is one channel against three, it receives a quarter of the objective, which is why AbsRel moves much less than the radiance metrics between the encoder and the latent-fit rows. Fitting on depth alone reaches an AbsRel of $0.019$, against $0.045$ for the joint fit and $0.049$ for the encoder, so the decoder can express better depth than the joint fit reports.

\vspace{-1em}
\paragraph{Synthetic scenes.}
We minimize $\mathcal{L}_{\text{optim}}$ with Adam for $500$ iterations at a learning rate of $0.1$ on $z$ and $s$, rendering the fitted view at $64$ samples per pixel. The image term uses $\lambda_{\log} = 0.1$, $\tau = 0.99$ and $c_{\max} = 10$; the repulsion term uses $\lambda_{\text{rep}} = 0.1$ and $m = 2$. Because the gradient is a Monte Carlo estimate, we use $\beta = (0.9, 0.99)$ and clip it to a global norm of $0.05$. The iterates are not monotone in the objective, so we keep the best iterate under $\mathcal{L}_{\text{optim}}$ rather than the last. The parameters are initialized as $z = 0$ and $e^{s} = 0.18$. 

\vspace{-1em}
\paragraph{Real images.}
Geometry is reconstructed as a triangulated unprojected depth estimated by MoGe-2~\cite{wang2025moge2} on the full image ($1024$px longest side) to preserve scene context. We then mask the mesh to the objects and shade it using the predicted normals. Rendering uses a $640$ px resolution at $128$ spp. Each connected component of the mask optimizes a sigmoid-parametrized albedo, metallic, and roughness (initialized at $0.5$). This setting is more ambiguous than the synthetic one, so we make several changes to the objective, all shared among the baselines where applicable:

\begin{itemize}
\item $1000$ steps at LR $0.1$, clipping latent step norms to $0.1$, with $\lambda_{\log} = 0.01$.

\item A Gaussian prior term regularizes latents toward the training manifold.

\item To leverage the unmasked room as partial observation, we constrain the estimated emitter mesh within the view frustum on the radiance and depth of the input observation.

\item We set $\lambda_{\text{rep}} = 1$ and include in the repulsion calculation points sampled within the view frustum, to prevent the emitter from occluding the scene objects. 

\item To avoid small rooms decoded from $z=0$, optimization initializes from a $\mathcal{N}(0, I)$ sample with larger geometry. The scene (camera and reference object points) is scaled and translated inside this mesh to improve the initial fit.

\end{itemize}

Probes in Fig.~\ref{fig:real} are mirror spheres placed on the object surface and rendered alongside it.

\subsection{Baselines details}

\paragraph{Direct optimization (Smoothness).}
Following DPI~\cite{lyu2023diffpostillum}, we use NeRFactor's smoothness prior~\cite{zhang2021nerfactor}, the summed squared one-pixel finite differences of a field $f$, with wrap-around on both axes,
\begin{equation}
    \mathcal{L}_{\text{smooth}}(f) = \sum_i (\Delta_u f)_i^2 + (\Delta_v f)_i^2 .
\end{equation}
NeRFactor penalizes linear radiance, but we observe better results in the log domain, and apply the same term to log depth in the 2.5D area light variant. The weights, $5 \times 10^{-5}$ on log radiance and $10^{-4}$ on log depth, were selected on a subset of mirror scenes by evaluation view PSNR. The fields are $128 \times 256$, matching the decoder's output.
Two variants of the same control isolate the role of the geometry, both optimizing the same radiance grid: an environment map at infinity, and a sphere of fixed radius per scene, the smallest origin-centered sphere enclosing every object of both views, enlarged by $20\%$.

\begin{figure}
    \centering
    \includegraphics[width=0.99\linewidth]{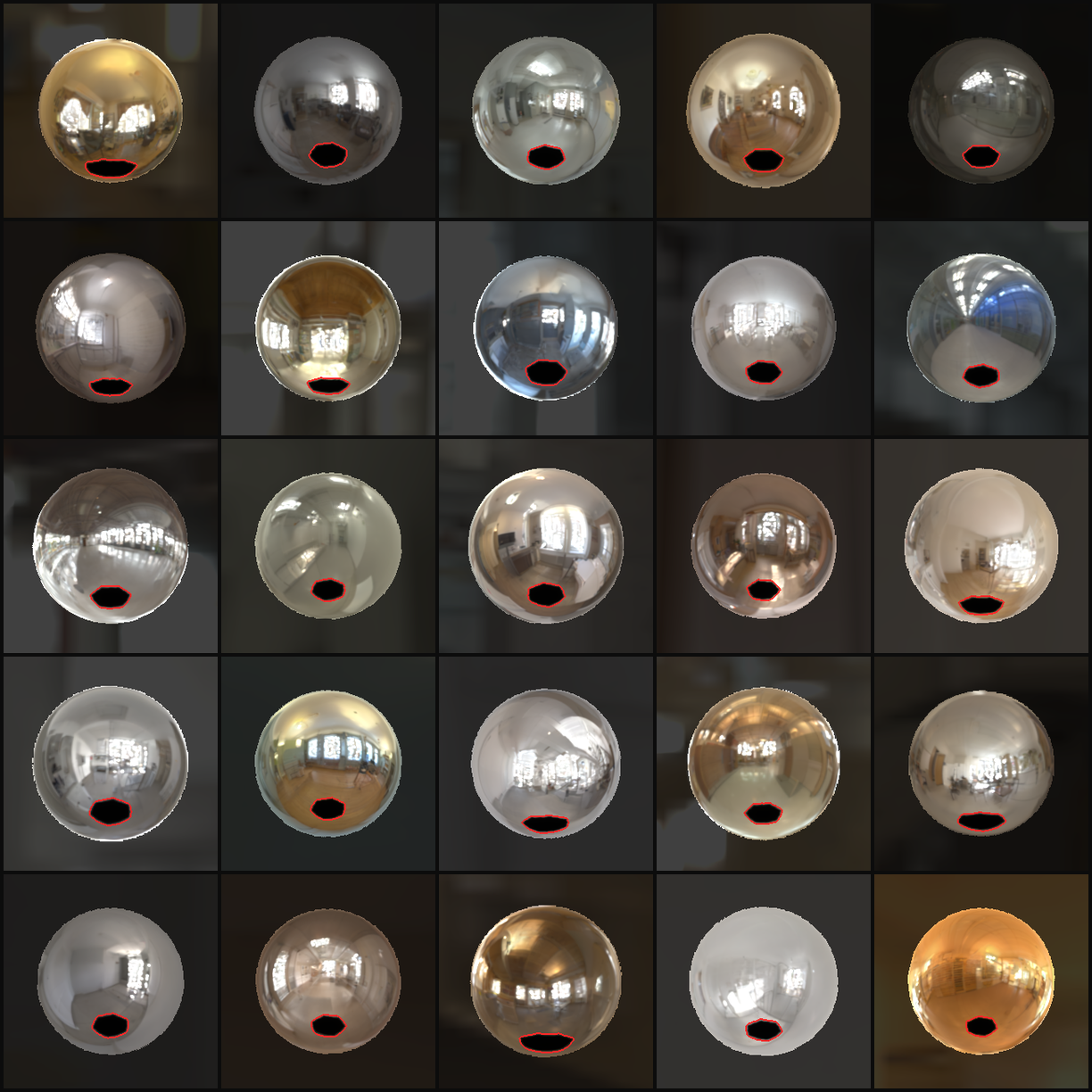}
    \caption{The nadir region DPI cannot represent. Each panel is the rendered evaluation mirror probe of one test scene under DPI's predicted environment.The dark patch low on each sphere is the part of the panorama DPI has no data for, and the red outline shows the mask that is used for computing the results in Tab. \ref{tab:eval_masked}.
    \vspace{-0.8em}}
    \label{fig:dpi_nadir}
\end{figure}

\vspace{-1em}
\paragraph{DPI.}
Retraining DPI~\cite{lyu2023diffpostillum} is reported to take one week on four A100 GPUs, so we run its released model with its own sampler and fitting procedure, bypassing only the material optimization. Its model was trained on the Laval Indoor HDR dataset~\cite{gardner2017indoor} without nadir inpainting, so every predicted panorama carries a black cap at the bottom. This highlights the importance of our preprocessing, but also entangles DPI's performance with a limitation of its training data.
To disentangle the two, we mask the probe pixels that DPI cannot represent (Fig.~\ref{fig:dpi_nadir}) and recompute the metrics on the rest, applying the same mask to every method and to the ground truth (Tab.~\ref{tab:eval_masked}). We report this only for the mirror probe, where each pixel reflects a single direction and the affected region is well defined; on the diffuse and matte probes the cap contributes to nearly every pixel. Masking improves DPI's scores and slightly lowers the others'. It substantially reduces DPI's deficit relative to VENI, but explains only a small part of its deficit relative to the 2.5D area light methods.

\begin{table}
\centering
\begin{adjustbox}{width=0.99\linewidth}
\small
\begin{tabular}{c l ccc}
\toprule 
& \multirow{3}{*}{Method} & \multicolumn{3}{c}{------------------Mirror------------------} \\
& & \textbf{PSNR} $\uparrow$ & \textbf{RMSE} $\downarrow$ & \textbf{ Angular Error} $\downarrow$ \\
& & LDR & HDR & RGB$^\circ$\\
\midrule
\multirow{4}{*}{\rotatebox[origin=c]{90}{ \footnotesize Full probe}}
& DPI \cite{lyu2023diffpostillum} & 15.61 & 2.845 & 5.758 \\
& VENI \cite{walker2026veni} & 17.32 & 2.766 & 5.289 \\
& VENI\textcolor{mygreen}{+D} & \colorbox{tabsecond}{21.20} & \colorbox{tabsecond}{2.363} & \colorbox{tabsecond}{4.840} \\
& \textcolor{mygreen}{Ours} & \colorbox{tabfirst}{22.85} & \colorbox{tabfirst}{2.150} & \colorbox{tabfirst}{4.372} \\
\midrule
\multirow{4}{*}{\rotatebox[origin=c]{90}{\footnotesize Masked nadir}}  
& DPI \cite{lyu2023diffpostillum} & 16.60 & 2.904 & 5.774 \\
& VENI \cite{walker2026veni} & 17.25 & 2.824 & 5.335 \\
& VENI\textcolor{mygreen}{+D} & \colorbox{tabsecond}{21.10} & \colorbox{tabsecond}{2.412} & \colorbox{tabsecond}{4.896} \\
& \textcolor{mygreen}{Ours} & \colorbox{tabfirst}{22.77} & \colorbox{tabfirst}{2.195} & \colorbox{tabfirst}{4.415} \\
\bottomrule
\end{tabular}
\end{adjustbox}
\caption{Inverse rendering on the mirror probe, with and without the region that DPI cannot represent. Upper block: evaluation on the full probe, as in Tab.~\ref{tab:reconstruction}. Lower block: the same fits, evaluated only on probe pixels outside DPI's nadir hole (mask contour shown in Fig.~\ref{fig:dpi_nadir}). Since VENI and Ours do not suffer from this limitation, masking favors DPI by excluding pixels on which it cannot produce valid predictions. \colorbox{tabfirst}{Best} and \colorbox{tabsecond}{second} per column.}
\vspace{-1.2em}
\label{tab:eval_masked}
\end{table}

\section{Additional results}
\label{sup:results}

\subsection{Fitting metrics}
Tab.~\ref{tab:reconstruction_fitted} reports the metrics of Tab.~\ref{tab:reconstruction} on the fitted view itself, i.e., on the bunny and armadillo the illumination was optimized against.
The environment-map methods do not merely generalize poorly to the held-out probe, but they fail to fit the observation in the first place.

\begin{table*}
\centering
\begin{adjustbox}{width=0.95\linewidth}
\small
\begin{tabular}{cc ccc ccc ccc}
\toprule 
\multirow{2}{*}{Method} &
\multirow{2}{*}{Representation}
& \multicolumn{3}{c}{\textbf{PSNR LDR} $\uparrow$} & \multicolumn{3}{c}{\textbf{RMSE HDR} $\downarrow$} & \multicolumn{3}{c}{\textbf{RGB Angular Error} $\downarrow$} \\
\cmidrule{3-5} \cmidrule{6-8} \cmidrule{9-11} 
 &  & Diffuse & Matte & Mirror & Diffuse & Matte & Mirror & Diffuse & Matte & Mirror \\
\midrule

 & Env map
& 29.00 & 24.59 & 21.61 & 0.069 & 0.409 & 1.172 & 1.069 & 2.346 & 3.099\\

 Smoothness & Sphere proxy
& 32.56 & 27.76 & 24.09 & 0.048 & 0.368 & 1.068 & 0.779 & 2.008 & \colorbox{tabsecond}{2.717}\\

 & \textcolor{mygreen}{2.5D AL}
& 33.99 & 30.34 & 26.25 & 0.042 & 0.315 & 0.991 & 0.678 & \colorbox{tabsecond}{1.808} & \colorbox{tabfirst}{2.481}\\

DPI~\cite{lyu2023diffpostillum} & Env map
& 26.12 & 22.46 & 20.06 & 0.086 & 0.556 & 1.361 & 1.396 & 3.236 & 4.279\\

VENI~\cite{walker2026veni} & Env map
& 30.61 & 25.26 & 21.63 & 0.061 & 0.363 & 1.154 & 1.113 & 2.736 & 4.088\\

VENI\textcolor{mygreen}{+D (ours)} & \textcolor{mygreen}{2.5D AL}
& \colorbox{tabsecond}{38.10} & \colorbox{tabsecond}{32.92} & \colorbox{tabsecond}{26.72} & \colorbox{tabsecond}{0.030} & \colorbox{tabsecond}{0.219} & \colorbox{tabsecond}{0.879} & \colorbox{tabsecond}{0.638} & 2.237 & 3.644\\

\textcolor{mygreen}{Ours} & \textcolor{mygreen}{2.5D AL}
& \colorbox{tabfirst}{40.00} & \colorbox{tabfirst}{35.47} & \colorbox{tabfirst}{29.70} & \colorbox{tabfirst}{0.025} & \colorbox{tabfirst}{0.127} & \colorbox{tabfirst}{0.675} & \colorbox{tabfirst}{0.523} & \colorbox{tabfirst}{1.764} & 3.153\\

\bottomrule
\end{tabular}
\end{adjustbox}
\caption{Quantitative evaluation on the fitted view. Metrics are computed on the input view the illumination was optimised against, over the pixels of its two objects, and averaged over 25 synthetic scenes. The 2.5D area light representation performs best by modeling spatially varying illumination. \colorbox{tabfirst}{Best} and \colorbox{tabsecond}{second best} results are highlighted.}
\vspace{-0.5em}
\label{tab:reconstruction_fitted}
\end{table*}

\subsection{Single-image illumination estimators}
\label{sup:single_image}

\begin{figure}
    \centering
    \includegraphics[width=0.99\linewidth]{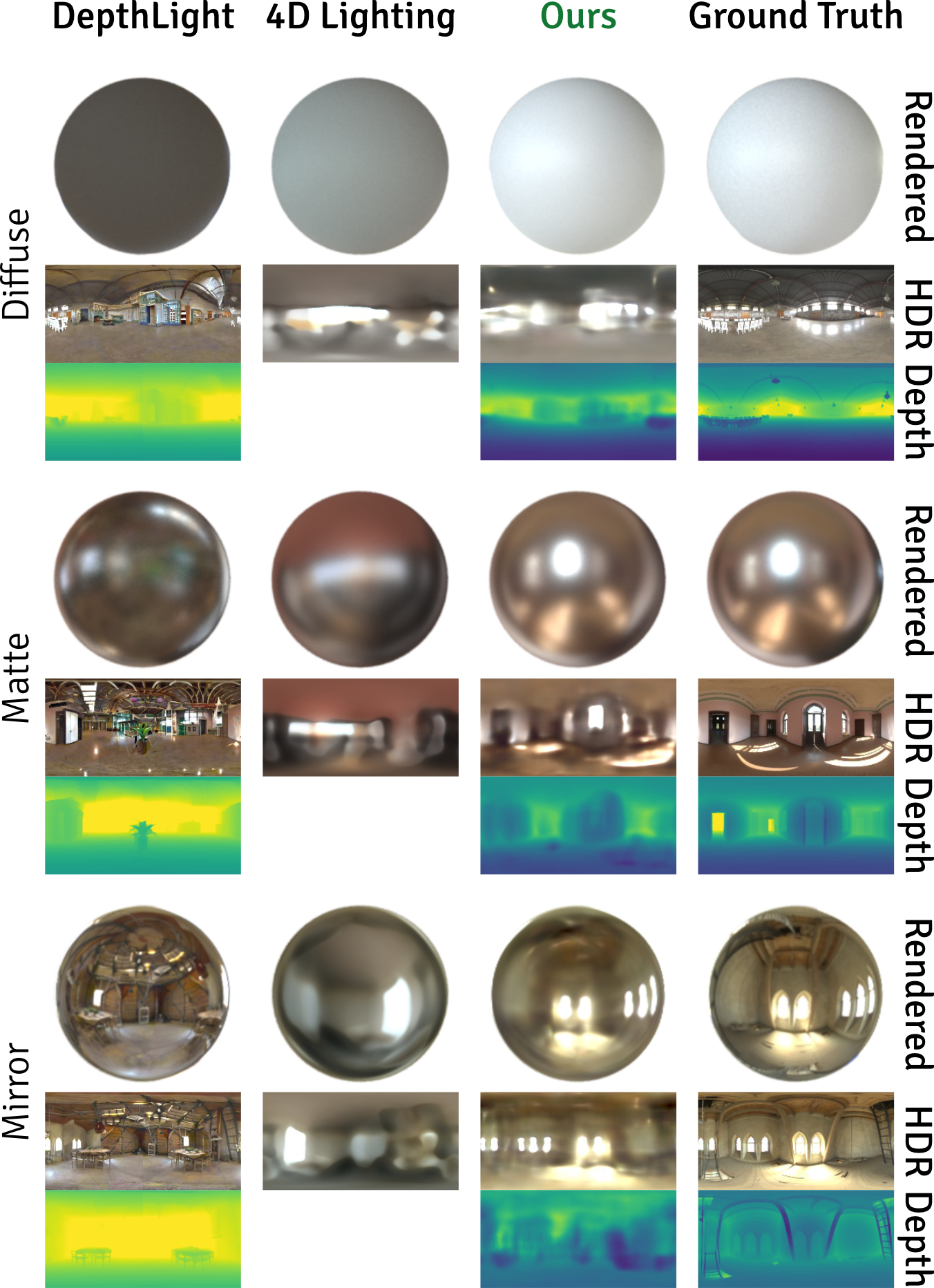}
    \caption{Complementary to Fig.~\ref{fig:results_synthetic}: the evaluation object re-rendered under the illumination predicted by DepthLight~\cite{manus2025depthlight} and 4D Lighting~\cite{tong2025spatio4dsig} from the fitted view, the predicted panorama and the depth where applicable.
    \vspace{-1em}}
    \label{fig:synth_others}
\end{figure}

Tab.~\ref{tab:reconstruction_others} and Fig.~\ref{fig:synth_others} report the two single-image estimators on the synthetic dataset. They are excluded from the main comparison because they predict illumination from the photograph alone, background included, whereas the optimization-based methods re-render known geometry and materials under a loss masked to the object. Moreover, 4D Lighting is built for video, and DepthLight for a wider field of view. Both are run unmodified.

\begin{table*}
\centering
\begin{tabular}{c ccc ccc ccc}
\toprule
\multirow{2}{*}{Method} & \multicolumn{3}{c}{\textbf{PSNR LDR} $\uparrow$} & \multicolumn{3}{c}{\textbf{RMSE HDR} $\downarrow$} & \multicolumn{3}{c}{\textbf{Angular Error} $\downarrow$} \\
\cmidrule{2-4} \cmidrule{5-7} \cmidrule{8-10}
 &  Diffuse & Matte & Mirror & Diffuse & Matte & Mirror & Diffuse & Matte & Mirror \\
\midrule

DepthLight~\cite{manus2025depthlight} 
& 9.67 & 11.18 & 12.14 & 0.366 & 0.918 & 3.034 & 8.653 & 7.731 & 11.375\\

4D Lighting~\cite{tong2025spatio4dsig}
& \colorbox{tabsecond}{14.83} & \colorbox{tabsecond}{14.13} & \colorbox{tabsecond}{13.51} & \colorbox{tabsecond}{0.306} & \colorbox{tabsecond}{0.878} & \colorbox{tabsecond}{2.896} & \colorbox{tabsecond}{6.455} & \colorbox{tabsecond}{6.949} & \colorbox{tabsecond}{7.931} \\

\textcolor{mygreen}{Ours}
& \colorbox{tabfirst}{33.66} & \colorbox{tabfirst}{30.35} & \colorbox{tabfirst}{22.85} & \colorbox{tabfirst}{0.089} & \colorbox{tabfirst}{0.271} & \colorbox{tabfirst}{2.178} & \colorbox{tabfirst}{1.189} & \colorbox{tabfirst}{2.275} & \colorbox{tabfirst}{4.378}\\

\bottomrule
\end{tabular}
\caption{Complementary to Tab.~\ref{tab:reconstruction}: the two single-image illumination estimators evaluated on the same 25 synthetic scenes and evaluation object, using identical metric definitions and scoring code. Their scores are not directly comparable to those of the illumination priors. DepthLight and 4D Lighting infer illumination from the photograph alone and use the entire image, including the background, whereas prior optimization is driven by re-rendering the scene’s known object geometry and materials and minimizes a loss masked to the object silhouette. \colorbox{tabfirst}{Best} and \colorbox{tabsecond}{second} per column.}
\vspace{-0.05in}
\label{tab:reconstruction_others}
\end{table*}

\subsection{Ablations}
\label{sec:exp:ablations}

Tab.~\ref{tab:ablation} ablates four components: the peak loss
$\mathcal{L}_\mathrm{peak}$, the Scharr gradient terms $\mathcal{L}^L_\nabla,
\mathcal{L}^D_\nabla$, decoder AdaGN, and explicit latitude channels $(\cos\theta, \sin\theta)$. Each is
removed individually and retrained under an otherwise identical configuration, then
evaluated both on latent fitting (Sec.~\ref{sec:exp:prior}) and on silver-matte
inverse rendering (Sec.~\ref{sec:exp:synth}), so that direct reconstruction quality
and fitness for downstream optimization are assessed separately.
Removing AdaGN costs the most in latent fitting, as it routes the latent code directly to every decoder scale rather
than forcing all global information through the convolutional stack. The explicit latitude channels behave in the opposite way: they do not affect the latent fitting yet their removal degrades all three
inverse-rendering metrics. Individually the effects are modest, but the components are complementary rather than
redundant. The fully ablated model is worse than the full
model on all eight reported metrics. Together the four address distinct failure modes
of a plain autoencoder trained on panoramas: clipped highlights, blurred structure,
weak global conditioning, and latitude-agnostic convolutions.

These four ablations vary the training objective and the decoder conditioning while holding the architecture fixed. VENI+D can be read as the coarsest ablation on this axis: it keeps our emitter, our fitting procedure and our data, and replaces the entire prior with a different architecture and training objective. Its gap to our model in Tab.~\ref{tab:prior_recon} and  Tab.~\ref{tab:reconstruction} is therefore attributable to the prior alone, and is larger than any single component removed here.

\begin{table}
\centering
\begin{adjustbox}{width=0.99\linewidth}
\small
\begin{tabular}{l cccc}
\toprule
Method & Params & Training & Decode & Fit per case \\
& (M) & (h) & (ms) & (min) \\
\midrule
DPI~\cite{lyu2023diffpostillum} & -- & $\approx$672 & -- & 8.4 \\
VENI~\cite{walker2026veni} & 25.2 & 6.5 & 32.7 & 2.1 \\
VENI\textcolor{mygreen}{+D} & 26.8 & 6.6 & 32.9 & 17.2 \\
\textcolor{mygreen}{Ours} & 13.9 & 2.1 & 4.3 & 9.9 \\
\bottomrule
\end{tabular}
\end{adjustbox}
\caption{Cost of the learned priors, measured on a RTX A5000, except the training time of DPI which is the one reported by its authors. Decode is one forward pass of the decoder at $128 \times 256$. Fit per case is one synthetic-scene optimization of $500$ iterations; DPI's fit uses a $1000$-step sampler.}
\vspace{-1em}
\label{tab:cost}
\end{table}

\subsection{Computational cost}
\label{sup:cost}

Tab.~\ref{tab:cost} summarizes the cost of the learned priors. During inverse rendering the decoder is not the bottleneck, as each iteration is dominated by path tracing. The gradients of a 2.5D area light, which flow through the emitter's geometry as well as its texture, are more expensive to compute than those of an environment map, which explains why fitting the observation through our prior takes significantly longer than through VENI.

\begin{table*}[t]
\centering
\begin{adjustbox}{width=0.95\linewidth}
\small
\begin{tabular}{l ccccc | ccc}
\toprule 
\multirow{3}{*}{Ablation} & \multicolumn{5}{c}{Latent Fitting} & \multicolumn{3}{c}{Inverse Rendering (Matte)} \\
& \textbf{PSNR} $\uparrow$ & \textbf{SSIM} $\uparrow$ & \textbf{LPIPS} $\downarrow$ & \textbf{PSNR} $\uparrow$ & \textbf{AbsRel} $\downarrow$ & \textbf{PSNR} $\uparrow$ & \textbf{RMSE} $\downarrow$ & \textbf{ Angular Error} $\downarrow$ \\
& \multicolumn{3}{c}{----------- \: LDR \: -----------}  &  HDR & Depth & LDR & HDR & RGB$^\circ$\\
\midrule
Ours & \colorbox{tabfirst}{25.34} & \colorbox{tabfirst}{0.751} & \colorbox{tabfirst}{0.422} & 37.12 & \colorbox{tabfirst}{0.045} & \colorbox{tabfirst}{30.41} & 0.272 & \colorbox{tabfirst}{2.269} \\
 w/o $\mathcal{L}_{\text{peak}}$ & 24.81 & 0.742 & 0.462 & 36.22 &	0.047 & 29.61 & 0.292 & 2.349 \\
  w/o $\mathcal{L}_{\nabla}^{L}$,  $\mathcal{L}_{\nabla}^{D}$ & 25.03 & 0.745 & 0.441 & 36.71 & 0.047 & 29.79 & 0.287 & 2.410 \\
  w/o AdaGN & 24.08 & 0.728 & 0.500 & 36.16 & 0.047 & 29.91 & \colorbox{tabfirst}{0.261} & 2.576 \\
  w/o $(\cos\theta, \sin\theta)$ channels & 25.26 & 0.749 & 0.428 & \colorbox{tabfirst}{37.20} & 0.046 & 29.87 & 0.279 & 2.404 \\
   w/o $\mathcal{L}_{\text{peak}}$, $\mathcal{L}_{\nabla}^{L}$,  $\mathcal{L}_{\nabla}^{D}$, AdaGN, $(\cos\theta, \sin\theta)$ & 22.94 & 0.708 & 0.566 & 34.93 & 0.050 & 29.35 & 0.322 & 2.714 \\
\bottomrule
\end{tabular}
\end{adjustbox}
\caption{Ablation of model components. Each component accounts individually for only a
small part of the performance, but the reconstruction degradations accumulate almost
additively, and the model with all four ablated is worse than the full model on every
metric.}
\vspace{-1em}
\label{tab:ablation}
\end{table*}

\end{document}